\documentclass[review]{elsarticle}
\usepackage[numbers]{natbib}

\usepackage{mathrsfs}
\usepackage{times}
\usepackage{bm}
\usepackage{amsfonts}
\usepackage{amsmath}
\usepackage{amssymb}
\usepackage{graphicx}
\usepackage{subfigure}
\usepackage{color}
\usepackage{algorithm}
\usepackage{algorithmic}
\usepackage{url}
\usepackage{verbatim}

\newtheorem{theorem}{Theorem}
\newtheorem{corollary}{Corollary}
\newtheorem{definition}{Definition}
\newtheorem{lemma}{Lemma}

\newcommand{\hint}[1]{}

\begin{document}

%---------------------------------FOR PATTERN RECOGNITION----------------------------------------------
%\begin{comment}
\begin{frontmatter}
\title{Relaxed Sparse Eigenvalue Conditions for Sparse Estimation via Non-convex Regularized Regression}
\author{Zheng Pan}
\ead{panz09@mails.tsinghua.edu.cn}
\author{Changshui Zhang\corref{cor:1}}
\ead{zcs@tsinghua.edu.cn}
\address{Department of Automation, Tsinghua University, Beijing 100084, P.R.China\\
State Key Lab of Intelligent Technologies and Systems\\
Tsinghua National Laboratory for Information Science and Technology(TNList)\\

}
\cortext[cor:1]{Corresponding author}
\begin{abstract}
Non-convex regularizers usually improve the performance of sparse estimation in practice. To prove this fact, we study the conditions of sparse estimations for the sharp concave regularizers which are a general family of non-convex regularizers including many existing regularizers. For the global solutions of the regularized regression, our sparse eigenvalue based conditions are weaker than that of L1-regularization for parameter estimation and sparseness estimation. For the approximate global and approximate stationary (AGAS) solutions, almost the same conditions are also enough. We show that the desired AGAS solutions can be obtained by coordinate descent (CD) based methods. Finally, we perform some experiments to show the performance of CD methods on giving AGAS solutions and the degree of weakness of the estimation conditions required by the sharp concave regularizers.
\end{abstract}
\end{frontmatter}

\section{Introduction}

High-dimensional estimation concerns the parameter estimation problems in which the dimensions of parameters are comparable to or larger than the sampling size. In general, high-dimensional estimation is ill-posed. Additional prior knowledge about the structure of the parameters is usually needed to obtain consistent estimations. In recent years, tremendous research works have demonstrated that the prior on \emph{sparsity} of the true parameters can lead to good estimators, e.g., the well-known work of compressed sensing \cite{candes2011probabilistic} and its extensions to general high-dimensional inference \cite{negahban2012unified}.

For high-dimensional sparse estimation, sparsity is usually imposed as sparsity-encouraging \cite{candes2008enhancing} regularizers for linear regression methods. Many regularizers have been proposed to describe the prior of sparsity, e.g., $\ell_0$-norm, $\ell_1$-norm, $\ell_q$-norm with $0<q<1$, smoothly clipped absolute deviation (SCAD) penalty \cite{fan2001variable}, log-sum penalty (LSP) \cite{candes2008enhancing}, minimax concave penalty (MCP) \cite{zhang2010nearly} and Geman penalty (GP) \cite{geman1995nonlinear, trzasko2009relaxed}. Except $\ell_1$-norm, all of these sparsity-encouraging regularizers are non-convex. Non-convex regularizers were proposed to improve the performance of sparse estimation in many applications, e.g., image inpainting and denoising \cite{shi2011nonconvex}, biological feature selection \cite{breheny2011coordinate, shen2012likelihood}, MRI \cite{candes2008enhancing, chartrand2009fast, trzasko2007sparse, trzasko2009highly, trzasko2008fixed} and CT \cite{ramirez2011nonconvex, sidky2007image}. However, it still lacks theoretical explanation for the improvement on sparse estimation for non-convex regularizers. This paper aims to establish such a theoretical analysis.

In the field of sparse estimation, the following three problems are typically studied. In this paper, we mainly study the first two problems.
\begin{enumerate}
\item Sparseness estimation: whether the estimation is as sparse as the true parameters;
\item Parameter estimation: whether the estimation is accurate in the sense that the error between the estimation and the true parameter is small under some metric;
\item Feature selection: whether the estimation correctly identities the non-zero components of the true parameters.
\end{enumerate}

For the sparseness estimation, the non-convex regularizers give better approximations to $\ell_0$-norm than the convex ones. They are more probable to encourage the regularized regression to yield sparser estimations than the convex regularizers. For example, $\ell_q$-regularization can give the sparsest consistent estimations even when $\ell_1$-regularization fails \cite{foucart2009sparsest}. However, $\ell_q$-norm has infinite derivatives at zero and zero vector is always a trivial local minimizer of the regularized regression. The non-convex regularizers with finite derivatives can remedy the numerical problem of $\ell_q$-norm, e.g., LSP, SCAD and MCP. These regularizers can also give sparser solutions for more general situations than $\ell_1$-regularization in experiments \cite{candes2008enhancing} and in theory \cite{fan2001variable, trzasko2009relaxed, zhang2010nearly, zhang2011general}.

For the parameter estimation, a lot of applications and experiments have demonstrated that many non-convex regularizers give good estimations with far less sampling sizes than $\ell_1$-norm as the regularizers \cite{candes2008enhancing, chartrand2009fast, ramirez2011nonconvex, sidky2007image, trzasko2007sparse, trzasko2009highly, trzasko2008fixed}. In theory, the requirements for the sampling sizes are essentially the requirements for design matrix or, rather, \emph{estimation conditions}. A weaker estimation condition means less sampling size needed or weaker requirements on design matrix. Weaker estimation conditions are important for the application in which the data dimension is very high while the sampling is expensive or restrictive. Theoretically, all of the non-convex regularizers mentioned above admit accurate parameter estimations under appropriate conditions, e.g., $\ell_q$-norm \cite{foucart2009sparsest}, MCP \cite{zhang2010nearly}, SCAD \cite{zhang2010nearly} and general non-convex regularizers \cite{zhang2011general}.

There are mainly two types of estimation conditions. The first is \emph{sparse eigenvalue} (SE) conditions, e.g., the restricted isometry property (RIP) \cite{candes2011probabilistic, candes2005decoding} and the SE used by \citet{foucart2009sparsest} and \citet{zhang2011sparse}. The second is \emph{restricted eigenvalue} (RE) conditions, e.g., the $\ell_2$-restricted eigenvalue ($\ell_2$-RE) \cite{bickel2009simultaneous, koltchinskii2009dantzig} and restricted invertibility factor (RIF) \cite{ye2010rate, zhang2011general}. Based on SE, \citet{foucart2009sparsest} gave a weaker estimation condition for $\ell_q$-norm than $\ell_1$-norm. \citet{trzasko2009relaxed} established a universal RIP condition for general non-convex regularizers including $\ell_1$-norm. Since the conditions proposed by \citet{trzasko2009relaxed} are regularizer-independent, it can not be weakened for non-convex regularizers unfortunately. The definition of SE is regularizer-independent while the RE is dependent on the regularizers. RE can give a regularizer-dependent estimation condition for general regularizers, e.g., the $\ell_2$-RE based work by \citet{negahban2012unified} and the RIF based work by \citet{zhang2011general}.

%Feature selection is another problem in which non-convex regularizers perform better than $\ell_1$-norm. Compared with non-convex regularizers, stronger conditions \cite{zhao2006model} are necessary for the $\ell_1$-regularized least squares estimation (LASSO) to achieve consistent feature selection \cite{tropp2006just, wainwright2009sharp, zhang2011general}. With consistent feature selection, we hope the estimation performs like a least squares estimation (LSE) with the support of the true parameters known, which is called \emph{oracle property}. However, LASSO fails to satisfy the oracle property even if the required stronger condition is provided \cite{fan2001variable, fan2004nonconcave}. Hence, non-convex regularizers are necessary to remedy the issue. Some weaker conditions of feature selection than LASSO have been proposed for SCAD \cite{fan2001variable}, $\ell_q$-norm \cite{zou2006adaptive, zou2008one, huang2008asymptotic}, MCP \cite{zhang2010nearly} and truncated $\ell_1$-norm \cite{shen2012likelihood, shen2013constrained}. \citet{zhang2011general} unified the conditions of these previous works for general non-convex regularizers.

However, the optimization for non-convex regularizers is difficult. It usually cannot be guaranteed to achieve a global optimum for general non-convex regularizers. Nevertheless, some optimization methods can lead to local optimums, e.g., coordinate descent \cite{breheny2011coordinate, mazumder2011sparsenet} and iterative reweighted (or majorization-minimization) methods \cite{candes2008enhancing, hunter2005variable, zhang2013multi, zhang2010analysis}, homotopy \cite{zhang2010nearly}, difference convex (DC) methods \cite{shen2012likelihood, shen2013constrained} and proximal methods \cite{gong2013general, pan2013high}. Hence, it is meaningful to analyze the performance of sparse estimation for these non-optimal optimization methods. For example, the multi-stage relaxation methods \cite{zhang2013multi, zhang2010analysis} and its one-stage version the adaptive LASSO \cite{huang2008adaptive, zou2006adaptive} replace the regularizers with their convex relaxations using majorization-minimization. Compared with LASSO, the multi-stage relaxation methods improve the performance on parameter estimation \cite{zhang2010analysis}. \citet{zhang2011general} use the solutions of LASSO as the initialization and continue to optimize by gradient descent. It is stated that LASSO followed by gradient descent can output an approximate global and stationary solution which is identical to the unique sparse local solution and the global solution. The multi-stage relaxation methods, the "LASSO + gradient descent" methods and the homotopy methods need the same SE or RE conditions as LASSO. The DC methods \cite{shen2013constrained} and the proximal methods \cite{pan2013high} need to know the sparseness of its solutions in advance to ensure the performance of parameter estimation, but these two methods cannot control the sparseness of its solutions explicitly.
%Before this paper, the existing theory still needs the same conditions as that of convex regularizers to predict the sparseness of a local minimizer in advance.

Based on the related work, we make the following contributions:

\begin{itemize}
\item
For a general family of non-convex regularizers, we propose new SE based estimation conditions which are weaker than that of $\ell_1$-norm. As far as we know, our estimation conditions are the weakest ones for general non-convex regularizers. The proposed conditions approach the SE conditions of $\ell_0$-regularized regression as the regularizers become closer and closer to $\ell_0$-norm. We also compare our SE conditions with RE conditions. For $\ell_1$-regularized regression, RE based estimation conditions are less severe than that based on SE \cite{bickel2009simultaneous}. However, for the case of non-convex regularizers, their relationship changes. For proper non-convex regularizers, SE conditions become weaker than RE conditions, because SE conditions can be greatly weakened from $\ell_1$-norm to non-convex regularizers while RE conditions remain the same.
\item
Under the proposed SE conditions, we establish upper bounds for the estimation error in $\ell_2$-norm. The error bounds are on the same order as that of $\ell_1$-regularized regression. It means that although the proposed SE conditions are weakened, the parameter estimation performance is not weakened. With appropriate additional conditions, we further give the results of sparseness estimations, which show the non-convex regularized regression give estimations with the sparseness on the same order as the true parameters.
\item
Like the global solutions of non-convex regularized regression, we show that the approximate global and approximate stationary (AGAS) solutions \cite{zhang2011general} also theoretically guarantee accurate parameter estimation and sparseness estimation. The error bounds of parameter estimation are on the order of noise level and the degrees of approximating the stationary solutions and the global optimums. If the degrees of these two approximations are comparable to the noise level, the theoretical performance on parameter estimation and sparseness estimation is also comparable to that of global solutions. Furthermore, the required estimation conditions are almost the same as that of global solutions, which means the estimation conditions for AGAS solutions are also weaker than that required by $\ell_1$-norm. The estimation result on AGAS solutions is useful for application since it shows the robustness of the non-convex regularized regression to the inaccuracy of the solutions and gives a theoretical guarantee for the numerical solutions.
\item
Under a mild SE condition, the approximate global (AG) solutions are obtainable and the approximation error is bounded by the prediction error. If the prediction error is small, the solution will be a good approximate global solution. The algorithms which control the sparseness of the solutions explicitly are suitable to give good AG solutions, e.g., OMP \cite{tropp2007signal} and GraDeS \cite{garg2009gradient}. For an AG solution, the coordinate descent (CD) methods update it to be approximate stationary (AS) without destroying its AG property. CD have been applied to regularized regression with
%convex regularizers \cite{friedman2010regularization, wu2008coordinate, friedman2007pathwise} and
non-convex regularizers \cite{breheny2011coordinate, mazumder2011sparsenet}. However, the previous works did not allow the non-convex regularizers to approximate $\ell_0$-norm arbitrarily. Our analysis does not have such restriction on the non-convex regularizers .
\end{itemize}

\textbf{Denotation.} We use $\bar{\mathcal T}$ to denote the complement of the set $\mathcal T$ and $|\mathcal T|$ to denote the number of elements in $\mathcal T$. For an index set $\mathcal T \in \{1, 2, \cdots, p\}$, $\theta_{\mathcal T}$ denotes the restriction of $\theta=(\theta_1, \theta_2, \cdots, \theta_p)$ on $\mathcal T$, i.e., $\theta_{\mathcal T} = (\theta_i: i\in \mathcal T)$. The support $\text{supp}(\theta)$ of a vector $\theta$ is defined as the index set composed of the non-zero components' indices of $\theta$, i.e., $\text{supp}(\theta)=\{i : \theta_i \not = 0\}$. The $\ell_0$-norm of the vector $\theta$ is the number of non-zero components of $\theta$, i.e., $\|\theta\|_0 =  |\text{supp}(\theta)|$.

\section{Preliminaries} \label{sec:5}
\hint{sec:5}

We first formulate the sparse estimation problems. Suppose we have $n$ samples $(y_1,z_1), (y_2,z_2), \cdots, (y_n,z_n) $, where $y_i \in \mathbb R$ and $z_i\in \mathbb R^p$ for $i=1,\cdots,n$. Let $X=(z_1, \cdots, z_n)^T \in \mathbb R^{n \times p}$ and $y=(y_1, \cdots, y_n)^T \in \mathbb R^{n}$. We assume there exists an $s$-sparse \emph{true parameter} $\theta^*$ which is supported on $\mathcal S$ and satisfies $y = X \theta^* + e$ with a small noise $e \in \mathbb R^n$. In this paper, we assume that the energy of the noise is limited by a known level $\epsilon$, i.e., $\|e\|_2 \le \epsilon$. For Gaussian noise $e \sim \mathcal N(0,\sigma I_n)$, this assumption is satisfied for $\epsilon = \sigma \sqrt{n+2\sqrt{n\log n}}$ with the probability at least $1-1/n$ \cite{cai2009recovery}.

We focus on using the following regularized regression to recover $\theta^*$ from $y$. This method uses the solutions of the following regularized regression as the estimations to the true parameters.
\hint{eqn:1}
\begin{equation}\label{eqn:1}
\hat \theta  = \arg\min_{\theta \in \mathbb R^p} \mathcal F(\theta),
\end{equation}
where $\mathcal F(\theta) = \mathcal L(\theta) + \mathcal R(\theta)$. $\mathcal L(\theta) = \|y - X \theta\|_2^2/(2n)$ is the \emph{prediction error}. $\mathcal R(\theta)$ is a non-convex regularizer. In this paper, we only study the \emph{component-decomposable} regularizer, i.e., $\mathcal R(\theta) = \sum_{i=1}^p r(|\theta_i|)$. We call $r(u)$ the \emph{basis function} of $\mathcal R(\theta)$. Table \ref{table:1} lists the basis functions of some popular regularizers. For the basis functions in Table \ref{table:1}, $r(u)$ has the formulation $r(u)=\lambda^2 r_0(u/\lambda;\gamma)$ where $r_0(u;\gamma)$ is a non-decreasing concave function over $[0,+\infty)$ and $\gamma$ is a parameter to describe the "degree of concavity", i.e., $r(u)$ changes from linear function of $u$ to the indicator function $I_{\{u \not = 0\}}$ as $\gamma$ varies from $+\infty$ to $0$ (except $\ell_1$-norm).

\begin{table}
  \centering
  \caption{Examples of popular regularizers. The second column is the basis functions of the regularizers. The third column is the zero gaps of the global solutions when the regularizers are $\xi$-sharp concave. The fourth column is the values of $\lambda^*$. Section \ref{sec:13} gives the proof for the result on $\lambda^*$ of LSP.}\label{table:1}
  \begin{tiny}
  \begin{tabular}{cccc}
  \hline
  Name & Basis Functions & Zero gap & $\lambda^*$\\

  \hline
  \hline

  $\ell_1$-norm & $r(u)=\lambda u$ & 0 & $\lambda^* = \lambda$ \\

  $\ell_q$-norm & $r(u)= \lambda^2 (u /\lambda)^q, \gamma = \log(1/(1-q))$ & $\lambda (q(1-q)/\xi)^{1/(2-q)}$ &  $\lambda^* = \lambda (2-q) \left( \frac{2(1-q)}{\xi} \right)^{\frac{q-1}{2-q}}$\\

  SCAD & $r(u)= \lambda \int_0^u \min \left\{1, \left(1- \frac{x/\lambda-1}{\gamma} \right)_+ \right\} dx$ & 0 &  $\lambda^* = \lambda$ for $\xi=1$\\

  LSP & $r(u)=\lambda^2 \log\left(1+ \frac{u}{\lambda \gamma}\right)$ &  $\max\{ \lambda (1/\sqrt{\xi} - \gamma),0\}$ &  $\lambda^* \le \lambda \sqrt{2 \xi \log (1+2/(\xi\gamma^2))}$\\

  MCP & $r(u)=\lambda \int_0^u \left(1- \frac{x}{\lambda \gamma}\right)_+ dx$ & $\lambda \sqrt{\gamma/\xi} $ &  $\lambda^* = \lambda \min\{\sqrt{\gamma \xi },1\}$\\

  GP & $r(u) = \lambda^2 u/(\lambda \gamma + u)$ & $\max\{ \lambda (\sqrt[3]{2\gamma /\xi} - \gamma),0\}$ &
  $\lambda^* = \left\{ \begin{array}{ll}
  \lambda(\sqrt{2\xi} - \xi \gamma/2), & \xi\gamma^2 \le 2\\
  \lambda/\gamma, & \xi\gamma^2 >2\\
  \end{array} \right. $\\
  \hline
  \end{tabular}
  \end{tiny}
\end{table}

Throughout this paper, we assume the basis function $r(u)$ satisfy the following properties. All of them hold for the basis functions in Table \ref{table:1}.

\begin{enumerate}
  \item $r(0)=0$;
  \item $r(u)$ is non-decreasing;
  \item $r(u)$ is concave over $[0,+\infty)$;
  \item $r(u)$ is continuous and piecewise differentiable. We use $\dot r(u+)$ and $\dot r(u-)$ to denote the right and left derivatives.
  \item $r(u)$ has the formulation $r(u) = \lambda^2 r_0(u/\lambda; \gamma)$, where $r_0(u;\gamma)$ is parameterized by $\gamma$ and is independent of $\lambda$.
\end{enumerate}

In this paper, the weaker SE based estimation conditions need two important properties: zero gap and null consistency \cite{zhang2011general}. Zero gap means the true parameters and the estimations are strong in the sense that the minimal magnitude of the non-zero components cannot be too close to zero. Null consistency requires that the regularized regression in Eqn. (\ref{eqn:1}) is able to identify the true parameter $\theta^*$ exactly when $\theta^* = 0$ and the error $e$ is inflated by a factor of $1/\eta>1$.

\begin{definition}[Zero Gap]\label{def:5}
\hint{def:5}
We say $\theta\in \mathbb R^p$ has a zero gap $u_0$ for some $u_0 \ge 0$ if $\min\{ |\theta_i|: i\in \text{supp}(\theta)\} \ge u_0$.
\end{definition}

\begin{definition}[Null Consistency]\label{def:6}
\hint{def:6}
Let $\eta\in(0,1)$. We say the regularized regression in Eqn. (\ref{eqn:1}) is $\eta$-null consistent if $
\min_{\theta} \|X \theta - e/\eta\|_2^2/(2n) + \mathcal R(\theta) = \|e/\eta\|_2^2/(2n)$.
\end{definition}

In order to guarantee the above two properties, we propose the following assumption, named \emph{sharp concavity}. Sharp concavity is important for our analysis because zero gap and null consistency can be derived from it.

\begin{definition}[Sharp Concavity]\label{def:4}
\hint{def:4}
We say a basis function $r(u)$ satisfies $C$-sharp concavity condition over an interval $\mathcal I$ if $r(u) > u \dot r(u-) +  C u^2/2 $
holds for any $u \in \mathcal I$, where $C$ is a positive constant. We also say $r(u)$ is $C$-sharp concave over $\mathcal I$ and a regularizer $\mathcal R(\theta)$ is $C$-sharp concave if its basis function is $C$-sharp concave.
\end{definition}

Strictly concave functions can only satisfy $r(u) > u \dot r(u-)$. However, if the left-derivative $\dot r(u-)$ decreases so fast that it admits a margin proportional to $u^2$ in some interval $\mathcal I$, the concave functions guarantee the sharp concavity.

$C$-sharp concavity is satisfied over $(0,u_0)$ if $r(u)$ is \emph{strongly concave} (or $-r(u)$ is strongly convex) over $(0,u_0)$ , i.e., for any $t_1,t_2 \in (0,u_0)$ and $\alpha\in [0,1]$,
\hint{eqn:57}
\begin{equation}\label{eqn:57}
r(\alpha t_1 + (1-\alpha) t_2) \ge \alpha r(t_1) + (1-\alpha) r(t_2) +  \frac{1}{2} C \alpha (1-\alpha) (t_1-t_2)^2.
\end{equation}
Section \ref{sec:14} shows that sharp concavity only needs Eqn. (\ref{eqn:57}) holds for $t_1=0$ and any $t_2\in(0,u_0)$, which means that the sharp concavity is weaker than the strong concavity. For example, MCP is  $((1+a)\gamma)^{-1}$-sharp concave over $(0,\sqrt{1+a}\lambda \gamma)$ for any $a>0$. Whereas, the strong concavity does not hold over $(\lambda \gamma, \sqrt{1+a}\lambda \gamma)$. Besides, $\ell_q$-norm holds $q(1-q)(u_0/\lambda)^{q-2}$-sharp concavity over $(0,u_0)$; LSP satisfies $\lambda^2/(\lambda \gamma +  u_0)^2$-sharp concavity over $(0, u_0)$; GP is $2 \lambda^3 \gamma / (\lambda \gamma + u_0)^3$-sharp concave over $(0, u_0)$.

Let $x_i$ be the i-th column of $X$ and
\[\xi = \max_{1\le i \le p} \|x_i\|_2^2/n.\]
We observe that $\xi$-sharp concavity derives non-trivial zero gaps and null consistency.

\begin{theorem}\label{thm:4}
\hint{thm:4}
If $r(u)$ is $\xi$-sharp concave over $(0,u_0)$, any global solution of Problem (\ref{eqn:1}) has a zero gap no less than $u_0$, i.e., $|\hat\theta_i| \ge u_0$ for any $i\in \mathrm{supp}(\hat\theta)$.
\end{theorem}

Table \ref{table:1} lists the zero gaps of $\hat\theta$ when the basis functions are $\xi$-sharp concave.
%Note that LSP and GP will lose their positive zero gaps for large $\gamma$ while MCP and $\ell_q$-norm always have positive zero gaps for any $\gamma>0$.

\begin{theorem}\label{thm:10}
\hint{thm:10}
Let $r(u)$ be $\xi$-sharp concave over $(0,u_0)$. The $\eta$-null consistency condition is satisfied if $r(u_0) \ge \frac{1}{2n\eta^2} \|e\|_2^2 $.
\end{theorem}

\citet{zhang2011general} give a probabilistic condition for null consistency when $X$ is drawn from Gaussian distributions. However, our condition is deterministic from the view of $X$. It is easy to check whether our condition holds. For the case of $r(u)=\lambda^2 r_0(u/\lambda;\gamma)$, the condition of Theorem \ref{thm:10} is $\lambda \ge \eta^{-1} b_0 \|e\|_2 / \sqrt{n}$, where $b_0 =1/\sqrt{2 r_0(u_0/\lambda;\gamma)}$ is a constant if $u_0 =O(\lambda)$ (all the regularizers in Table \ref{table:1} satisfy $u_0 = O(\lambda)$). Hence, we assume
\hint{eqn:20}
\begin{equation}\label{eqn:20}
\lambda = \eta^{-1} b_0 \epsilon/\sqrt{n}
\end{equation}
in this paper, so that the $\eta$-null consistency holds. In addition, we define
\hint{eqn:15}
\begin{equation}\label{eqn:15}
\lambda^* = \inf_{u>0}\{\xi u/2 + r(u)/u\}.
\end{equation}
$\lambda^*$ provides a natural normalization of $\lambda$ \cite{zhang2011general}. Table \ref{table:1} lists the values of $\lambda^*$ of the regularizers.
We observe $\lambda^* = O(\lambda)$ from Table \ref{table:1}. In general, for $r(u)=\lambda^2 r_0(u/\lambda;\gamma)$, we can define a constant $a_\gamma$ (independent to $\lambda$),
\hint{eqn:21}
\begin{equation}\label{eqn:21}
a_\gamma = \inf_{u>0} \{\xi u/2 + r_0(u;\gamma)/u\},
\end{equation}
so that $\lambda^* = a_\gamma \lambda $. Thus, we have
\hint{eqn:58}
\begin{equation}\label{eqn:58}
\lambda^* = \eta^{-1}a_\gamma b_0 \epsilon/\sqrt{n}.
\end{equation}

If the basis function $r(u)$ is linear over $(0, u)$ for some $u>0$, it is not sharp concave, e.g., SCAD and truncated $\ell_1$-norm \cite{zhang2010analysis}. We name such regularizers that are linear near the origin as \emph{weak non-convex regularizers}. The zero gaps of the global solutions with such regularizers cannot be guaranteed to be strictly positive.

%However, with extra conditions, our results on sparse estimation are also applicable for the weaker non-convex regularizers without sharp concavity. Therefore, we will mention explicitly when the sharp concavity is needed.

\section{Sparse Estimation of Global Solutions}\label{sec:1}\hint{sec:1}

In this section, we show our results on the SE based sparse estimation.

\begin{definition}[Sparse Eigenvalue]
For an integer $t\ge 1$, we say that $\kappa_-(t)$ and $\kappa_+(t)$ are the minimum and maximum sparse eigenvalues(SE) of a matrix $X$ if
\hint{eqn:51}
\begin{equation}\label{eqn:51}
\kappa_-(t) \le \frac{\|X\Delta\|_2^2}{ n\|\Delta\|_2^2} \le \kappa_+(t) \text{ for any } \Delta \text{ with } \|\Delta\|_0 \le t.
\end{equation}
\end{definition}

The SE is related to the restricted isometry constant (RIC) $\delta_{t}$ \cite{candes2011probabilistic, candes2005decoding}, which satisfies $1-\delta_t \le \|X\Delta\|_2^2/(n\|\Delta\|_2^2) \le 1+\delta_t$ for all $\Delta$ with $\|\Delta\|_0 \le t$. Thus, it follows that $\delta_t = (\kappa_+(t) - \kappa_-(t)) / (\kappa_+(t) + \kappa_-(t))$, where $\delta_t$ is actually the RIC of the scaled matrix $2 X / (\kappa_+(t) + \kappa_-(t))$. We employ SE since it allows $\kappa_+(t) \ge 2$ and avoids the scaling problem of RIC \cite{foucart2009sparsest}.

\begin{figure}
  \centering
  \subfigure[~]{\includegraphics[width=0.32\textwidth]{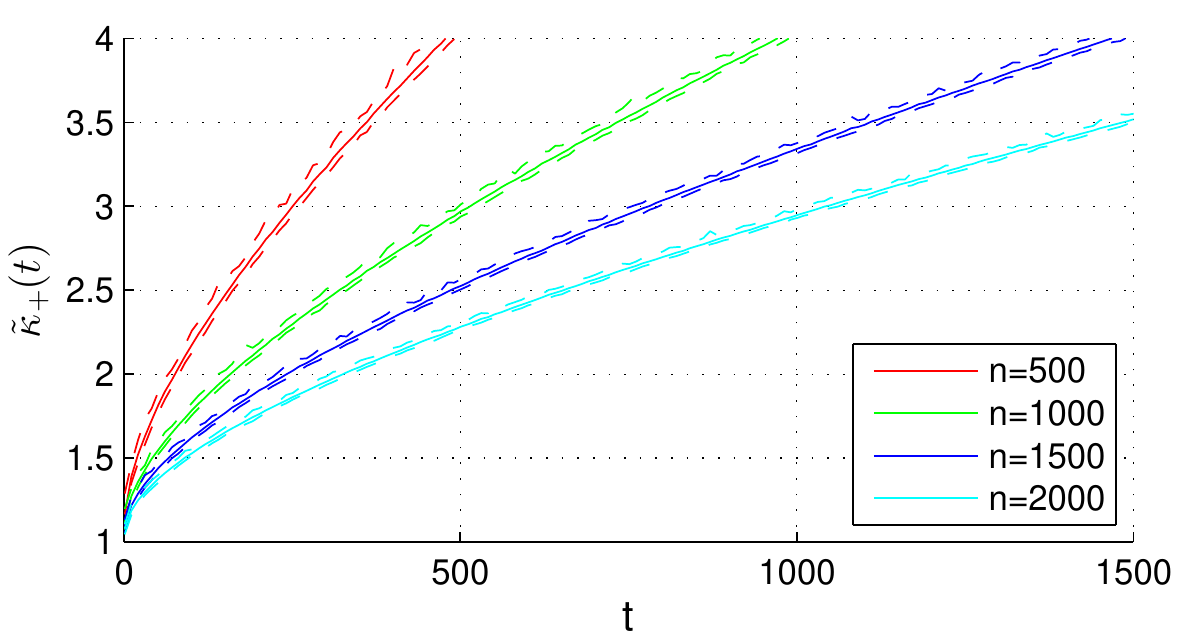}\label{fig:4a}}
  \subfigure[~]{\includegraphics[width=0.32\textwidth]{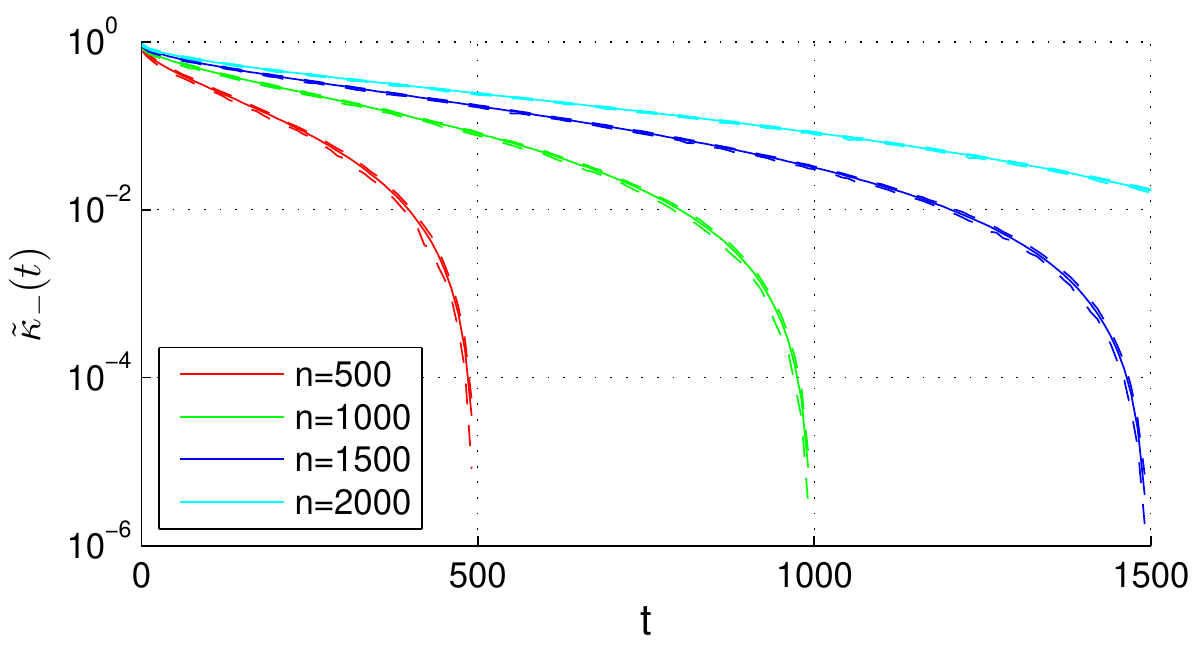}\label{fig:4b}}
  \subfigure[~]{\includegraphics[width=0.32\textwidth]{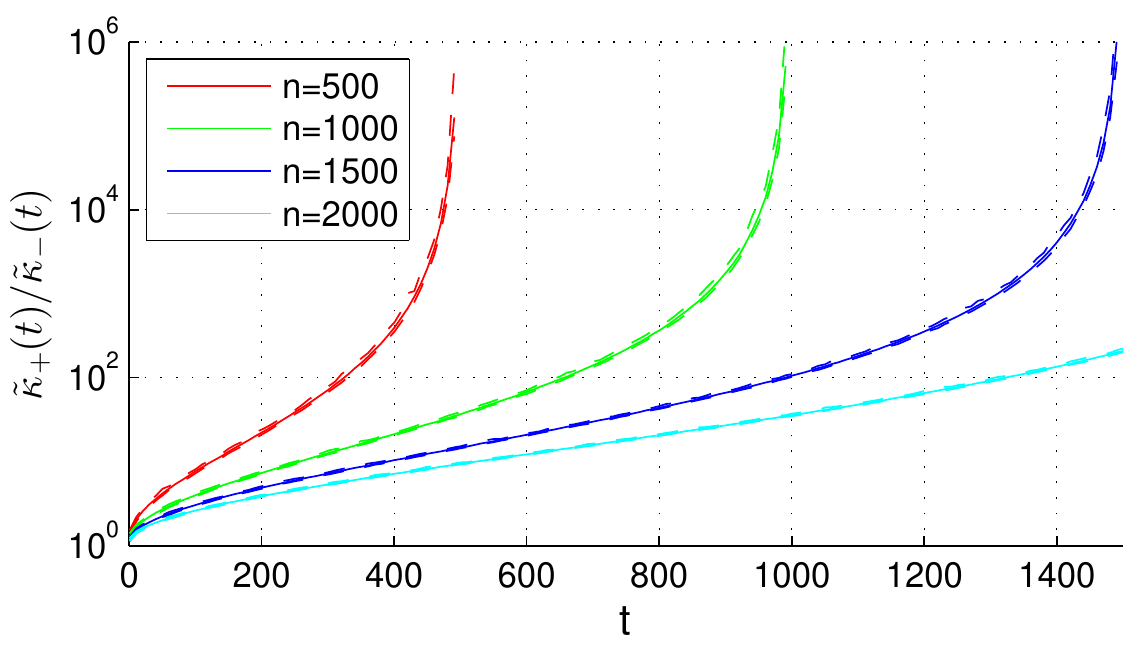}\label{fig:4c}}
  \caption{$\tilde\kappa_+(t)$, $\tilde\kappa_-(t)$ and $\tilde\kappa_+(t)/\tilde\kappa_-(t)$ for the Gaussian random matrices with $p=10~000$, $n=500, 1000, 1500, 2000$ and $t$ ranges from 1 to $n$. The solid lines are the average values of the 100 trials and the two dash lines around each solid line are the maximum and the minimum of the 100 trials.}
  \label{fig:4}
\end{figure}

In order to show the typical values of $\kappa_+(t)$ and $\kappa_-(t)$, we compute them and their ratio $\kappa_+(t)/\kappa_-(t)$ for the standard Gaussian $n \times p$ matrix\footnote{The elements are i.i.d. drawn from the standard Gaussian distribution $\mathcal N(0,1)$.}, where we fix $p=10~000$, $n=500$, $1000$, $1500$, $2000$ and $t$ varies from $1$ to $n$. It should be noted that $\kappa_+(t)$ and $\kappa_-(t)$ cannot be obtained efficiently. We use the following approximation method: For a matrix $X \in \mathbb R^{n\times p}$, we randomly sample its 100 submatrices $X_1, X_2, \cdots, X_{100} \in \mathbb R^{n\times t}$ composed by $t$ columns of $X$ and regard $\tilde\kappa_+(t) = \max_i \lambda_{\max}(X_i^T X_i)$ and $\tilde\kappa_-(t) = \min_i \lambda_{\min} (X_i^T X_i)$ as the approximations for $\kappa_+(t)$ and $\kappa_-(t)$, where $\lambda_{\max}(A)$ and $\lambda_{\min}(A)$ mean the maximal and minimal eigenvalues of $A$. Actually, $\tilde\kappa_+(t) \le \kappa_+(t)$ and $\tilde\kappa_-(t) \ge \kappa_-(t)$. For each $n$ and $t$, we generate 100 standard Gaussian matrices and compute the maximums, minimums and the means of the values of $\tilde\kappa_+(t)$, $\tilde\kappa_-(t)$ and $\tilde\kappa_+(t)/\tilde\kappa_-(t)$ for the 100 trials. Figure \ref{fig:4} illustrates the results. The variances of $\tilde\kappa_+(t)$, $\tilde\kappa_-(t)$ and $\tilde\kappa_+(t)/\tilde\kappa_-(t)$ with the same $n$ and $t$ are small since the corresponding lines for the maximum, minimum and mean values are close to each other. However, $\tilde\kappa_+(t)/\tilde\kappa_-(t)$ grows very fast as $t$ grows or $n$ decreases.

%For an example with $t=200$, $\tilde\kappa_+(t)/\tilde\kappa_-(t) = 22.4,~ 7.36,~ 4.90$ and $3.88$ for $n=500, 1000, 1500$ and $2000$ respectively. For another example with $t=400$, $\tilde\kappa_+(t)/\tilde\kappa_-(t) = 400.0$, $21.2$, $10.4$ and $7.2$ for $n=500, 1000, 1500$ and $2000$ respectively.

Based on SE, we establish the following parameter estimation result for global solutions of non-convex regularized regression. Let $\hat \rho_0$ and $\rho_0^*$ be the zero gaps of the global solution $\hat\theta$ and the true parameter $\theta^*$ respectively. Denote
\hint{eqn:60}
\begin{equation}\label{eqn:60}
\rho_0 =\min\{ \hat \rho_0, \rho^* \}.
\end{equation}

\begin{theorem}[Parameter Estimation of Global Solutions]\label{thm:1}
\hint{thm:1}
Suppose the following conditions hold.
\begin{enumerate}
\item $r(u)$ is invertible for $u\ge 0$ and $ r^{-1}(u/s_1) / r^{-1}(u/s_2)$ is a non-decreasing function of $u$ for any $s_2 \ge s_1 \ge 1$;
\item The regularized regression satisfies $\eta$-null consistency;
\item The following SE condition holds for some integer $t \ge \alpha s$,
\hint{eqn:5}
    \begin{equation}\label{eqn:5}
    \kappa_+(2t)/ \kappa_-(2t) < 4(\sqrt{2}-1) H_r(\rho_0, \alpha, s, t) +1,
    \end{equation}
    where $s=\|\theta^*\|_0$, $\alpha = \frac{1+\eta}{ 1- \eta}$, $H_r(\rho_0, \alpha, s, t) = \sqrt{ \frac{s}{t} } \frac{r^{-1}( r(\rho_0)/s)}{ r^{-1}(\alpha r(\rho_0)/t)}$ for $\rho_0>0$ and $H_r(0, \alpha, s, t) = \lim_{\rho\to 0+} H_r(\rho, \alpha, s, t)$.
\end{enumerate}
Then,
\hint{eqn:9}
\begin{equation}\label{eqn:9}
\|\hat\theta - \theta^*\|_2 \le C_1 \lambda^*,
\end{equation}
where $C_1 = \frac{(1+\sqrt{2})(1+\eta)\sqrt{t}}{\kappa_-(2t)}  \frac{H_r(\rho_0, \alpha, s, t) + 1/2}{H_r(\rho_0, \alpha, s, t) - (1+\sqrt{2})(\kappa_+(2t) / \kappa_-(2t) - 1)/4}$.
\end{theorem}

Since $\lambda^*$ is on the order of noise level $\epsilon$ (Eqn. (\ref{eqn:58})), the estimation error $\|\hat\theta - \theta^*\|_2$ is at most on the order of noise level. We give a detailed discussion on Theorem \ref{thm:1} in Section \ref{sec:12}. Before the discussion, we first show a corollary given in Section \ref{sec:12}, which shows that our SE condition only needs $\kappa_-(t) >0$ with $t=O(s)$. This SE condition is much weaker than that of $\ell_1$-norm. In fact, it is almost optimal since it is the same as the estimation condition of $\ell_0$-regularization \cite{foucart2009sparsest, zhang2011general}.

\begin{corollary}\label{coro:1}\hint{coro:1}
Let the condition 1 and 2 of Theorem \ref{thm:1} hold and $H_r(\rho_0,\alpha,s,\alpha s+1) \to \infty$ as $\gamma\to 0$. If $\kappa_-(2\alpha s +2) >0$, there exists $\gamma>0$ such that $\|\hat\theta - \theta^*\|_2 \le O(\lambda^*)$.
\end{corollary}

In addition to the error bound in Theorem \ref{thm:1}, we hope that the regularized regressions yield enough sparse solutions.
%Thus, estimating the sparseness of the global solutions is also an important problem.
We extend the results from \citet{zhang2011general} and show that the global solutions are sparse under appropriate conditions.

\begin{theorem}[Sparseness Estimation of Global Solutions]\label{thm:8}
\hint{thm:8}
Suppose the conditions of Theorem \ref{thm:1} hold. Consider $l_0>0$ and integer $m_0>0$ such that $$\sqrt{ 2 t \kappa_+(m_0) r(C_2 (1+\eta) \lambda^*)/m_0 } + \|X^T e/n\|_\infty < \dot r(l_0-),$$ where $C_2$ is defined in Eqn. (\ref{eqn:13}). Then, $|\text{supp}(\hat\theta) \backslash \mathcal S| \le m_0 +  t r( C_2 (1+\eta) \lambda^* )/ r(l_0)$.
\end{theorem}

\begin{corollary}\label{coro:2}
\hint{coro:2}
Suppose the basis function $r(u) = \lambda^2 r_0(u/\lambda)$ and the conditions of Theorem \ref{thm:8} hold with $t=(\alpha+1)s$, $m_0=\beta_0 s$ and $l_0 = \beta_1 \lambda$ for some $\beta_0,~\beta_1>0$. Let $C_3 = C_2 (1+\eta)a_\gamma $ where $C_2$ is the same as Theorem \ref{thm:8} and $a_\gamma$ is defined in Eqn. (\ref{eqn:21}). If
\hint{eqn:32}
\begin{equation}\label{eqn:32}
\frac{2(\alpha+1) \kappa_+(\beta_0 s)}{\beta_0} < \frac{(\dot r_0(\beta_1 -) - \eta a_\gamma)^2}{r_0(C_3)},
\end{equation}
then
\hint{eqn:33}
\begin{equation}\label{eqn:33}
|\text{supp}(\hat\theta) \backslash \mathcal S| \le (\beta_0 + (\alpha+1) r_0(C_3)/r_0(\beta_1)) s.
\end{equation}
\end{corollary}

\textbf{Example for Corollary \ref{coro:2}.} Consider the example of LSP with $r_0(u) = \log(1+u/\gamma)$ and $\beta_1=\sqrt{\gamma}$. Suppose the columns of $X$ are normalized so that $\xi=1$. Section \ref{sec:13} shows that $a_\gamma \le \sqrt{2 \log (1+2/\gamma^2)}$. Thus, the right hand of Eqn (\ref{eqn:32}) is larger than
\[
\frac{\left( 1/(1+\sqrt{\gamma}) - \eta \sqrt{2 \gamma \log(1+2/\gamma^2)} \right)^2}{ \gamma \log ( 1+ \gamma^{-1} C_2(1+\eta) \sqrt{2 \log(1 + 2/\gamma^2) } )}
\]
Thus, as $\gamma$ goes to 0, the right side of Eqn. (\ref{eqn:32}) is arbitrarily large. Eqn. (\ref{eqn:32}) holds for enough small $\gamma$. The right side of Eqn. (\ref{eqn:33}) is $\beta_0 s + O(s)$ as $\gamma \to 0$. Hence, we can freely select $\beta_0$ satisfying Eqn. (\ref{eqn:32}) with enough small $\gamma$. For example, if $\beta_0=1/s$, Eqn. (\ref{eqn:32}) holds for enough small $\gamma$ and Eqn. (\ref{eqn:33}) becomes
%\hint{eqn:46}
\begin{equation}\label{eqn:46}
|\text{supp}(\hat\theta)/\mathcal S| \le 1 + s(\alpha+1) \frac{ \log ( 1+ \gamma^{-1} C_2(1+\eta) \sqrt{2 \log(1 + 2/\gamma^2) } )}{ \log (1 + 1/\sqrt{\gamma})}.
\end{equation}
The right side of Eqn (\ref{eqn:46}) is at most on the order of $s$ when $\gamma$ is close to zero.

\section{Discussion on Theorem \ref{thm:1}}\label{sec:12}\hint{sec:12}

This section gives some detailed discussion on Theorem \ref{thm:1}.

\subsection{Invertible approximate regularizers}

If $r(u)$ is not invertible, e.g., MCP, we can design invertible basis function to approximate it. For example, we can use the following invertible function, named \emph{Approximate MCP}, to approximate MCP.
\hint{eqn:26}
\begin{equation}\label{eqn:26}
r(u) = \left\{\begin{array}{ll}
\lambda u - u^2/(2\gamma), & 0 \le u \le \lambda \gamma (1-\phi),\\
\frac{1}{2} \lambda^2 \gamma (1-\phi^2) \left( \frac{u}{\lambda\gamma (1-\phi)} \right)^{2\phi/(1+\phi)}, & u > \lambda \gamma(1-\phi),
\end{array}\right.
\end{equation}
where $\phi\in (0,1)$.
Approximate MCP is concatenated by the part of MCP over $[0,\lambda\gamma(1-\phi)]$ and the part of $\ell_{q}$-norm over $(\lambda\gamma(1-\phi), \infty)$ with $q = 2\phi/(1+\phi)$.
When $\phi \to 0$, $r(u)$ will become the basis function of MCP. We will address the method to obtain Eqn. (\ref{eqn:26}) in Section \ref{sec:8}. Any other non-invertible regularizers in Table \ref{table:1} can be approximated in the same way.

\subsection{Non-decreasing property of $ r^{-1}(u/s_1) / r^{-1}(u/s_2)$}

It can be verified that all the regularizers  in Table \ref{table:1} or their invertible approximate ones (in the way of Eqn. (\ref{eqn:26})) satisfy the non-decreasing property of $\frac{ r^{-1}(u/s_1) }{ r^{-1}(u/s_2)}$ for any $s_2\ge s_1 >0$. In fact, for derivative basis functions, this non-decreasing property is equal to that $u\dot r(u)/r(u)$ is a non-increasing function of $u$.
%We derive from the $C$-sharp concavity that $u \dot r(u)/r(u)$ is upper bounded by $1 - \frac{C}{2} \frac{u^2}{r(u)}$. The upper bound is a non-increasing function of $u$. Hence, sharp concave functions are very likely to satisfy the non-decreasing property of $\frac{ r^{-1}(x/s_1) }{ r^{-1}(x/s_2)}$.

\subsection{Non-sharp concave regularizers}

If $r(u)$ is not $\xi$-sharp concave, e.g., SCAD or LSP with $\gamma^2 > 1/\xi$, we cannot guarantee $\hat\theta$ has a positive zero gap.
%The condition 1 of Theorem \ref{thm:1} still holds for these regularizers or their approximate regularizers.
In this case, the condition 2 (null consistency) of Theorem \ref{thm:1} can be guaranteed by the $\ell_2$-regularity conditions \cite{zhang2011general} and the condition 3 becomes $ \kappa_+(2\alpha s)/ \kappa_-(2 \alpha s) <  1.65/\sqrt{\alpha} +1$ with $t=\alpha s$, which also belongs to the $\ell_2$-regularity conditions.
Hence, without $\xi$-sharp concavity, Theorem \ref{thm:1} still holds. Intuitively, non-sharp concave regularizers need the same estimation conditions as $\ell_1$-regularization since they cannot approximate $\ell_0$-norm arbitrarily.

\subsection{Relaxed SE based estimation conditions}

Much more relaxed estimation conditions are sufficient for $\xi$-sharp concave regularizers. Suppose $r(u)$ is $\xi$-sharp concave over $(0,\rho_0)$ with $0< \rho_0 \le \min_{i \in \mathcal S} |\theta^*_i|$. In this case, $H_r(\rho_0,\alpha, s,t )$ can become arbitrarily large for proper regularizers so that the SE condition in Eqn. (\ref{eqn:5}) is much weaker than the SE conditions of $\ell_1$-regularized regression. We have shown in Figure \ref{fig:4} that  $\tilde\kappa_+(t)/\tilde\kappa_-(t)$ ($\le \kappa_+(t)/\kappa_-(t)$) increases very fast as $t$ increases or $n$ decreases. Thus, a weaker constraint on $\kappa_+(2t) / \kappa_-(2t)$ in Eqn. (\ref{eqn:5}) is very important for sparse estimation problems.

\begin{figure}
  \centering
  \subfigure[Approximate MCP vs. L1]{\includegraphics[width=0.3\textwidth]{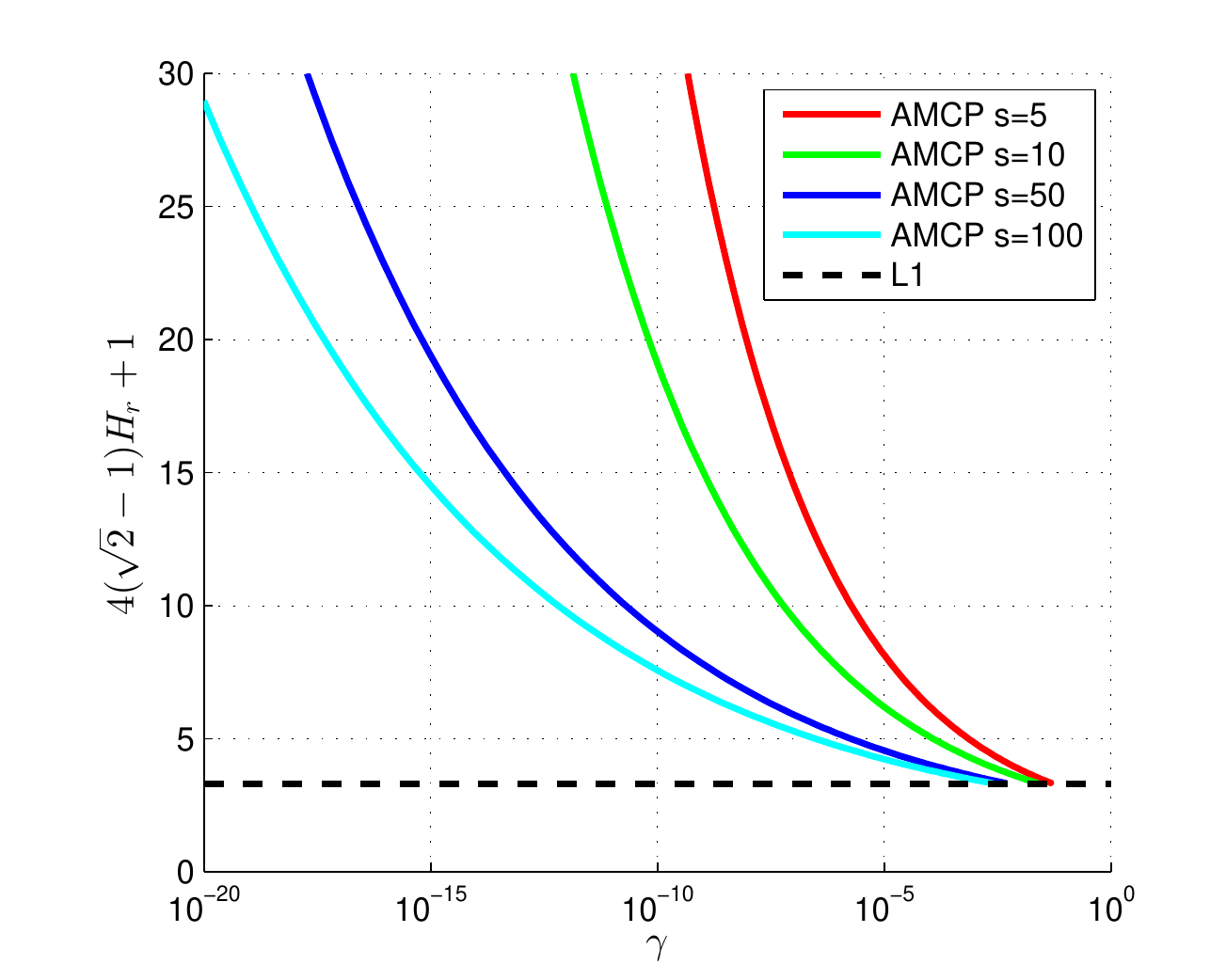}}
  \subfigure[LSP vs. L1]{\includegraphics[width=0.3\textwidth]{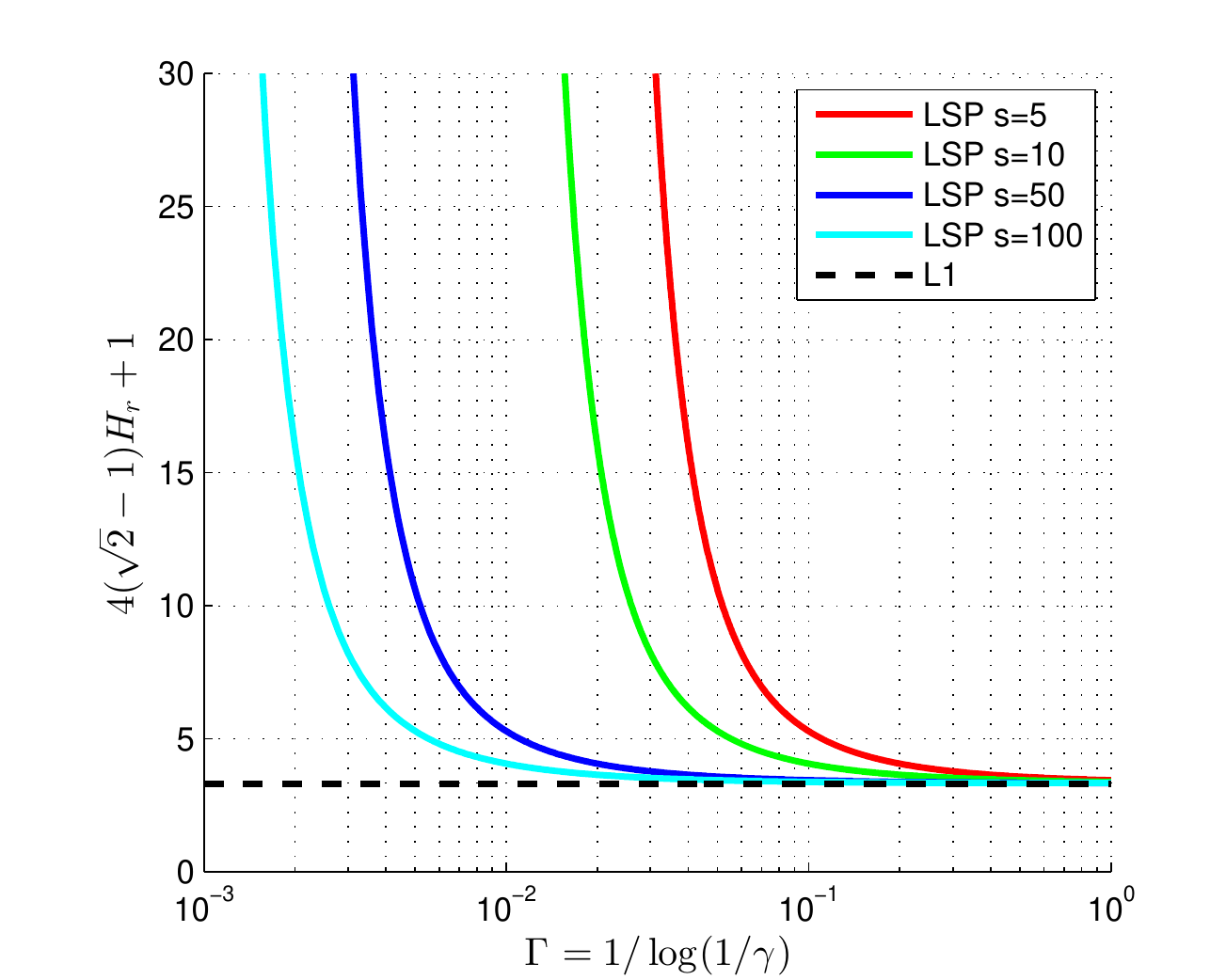}}
  \subfigure[$\ell_q$-norm vs. L1]{\includegraphics[width=0.3\textwidth]{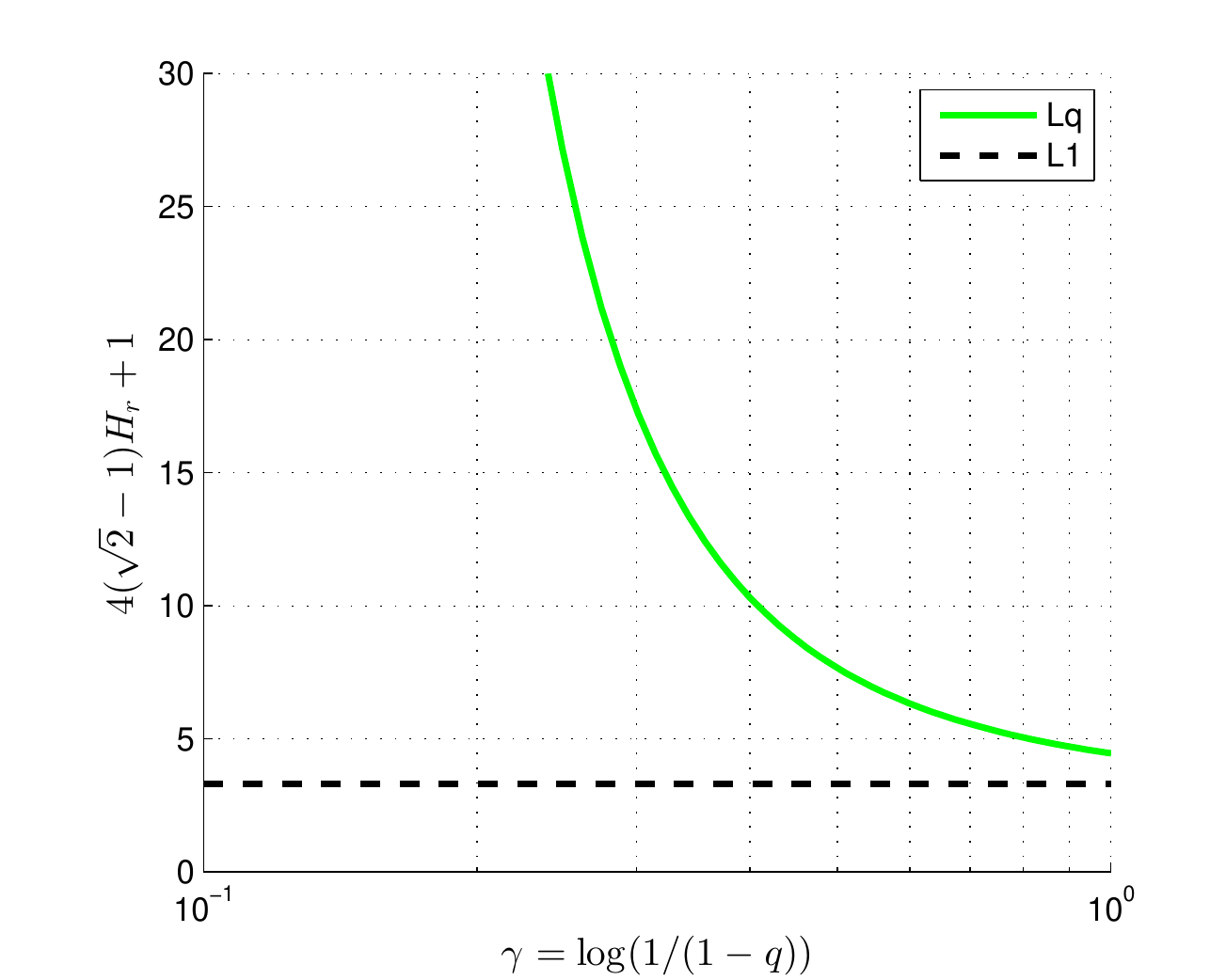}}
  \caption{\hint{fig:5}The upper bounds of the SE conditions for LSP, approximate MCP(AMCP) and $\ell_q$-norm. We set $\alpha=1.01$, $t=2s$. In each subfigure, we also plot the upper bound of SE conditions for $\ell_1$-norm, i.e., the right hand of Eqn. (\ref{eqn:45}) with $q=1$.}
  \label{fig:5}
\end{figure}

Here, we give the examples of approximate MCP, $\ell_q$-norm and LSP. For approximate MCP, Eqn. (\ref{eqn:16}) gives its $H_r(\rho_0,\alpha,s,t)$ (see Section \ref{sec:8}).
\hint{eqn:16}
\begin{equation}\label{eqn:16}
H_r(\rho_0, \alpha, s, t) = \alpha^{-1/2} (t/(\alpha s))^{\frac{1}{2\phi}},
\end{equation}
where we set $\gamma \xi = \frac{\phi}{1+\phi} (\alpha/t)^{1/\phi}$. For $\ell_q$-norm, the SE conditions can be written as
\hint{eqn:45}
\begin{equation}\label{eqn:45}
\frac{\kappa_+(2t)}{\kappa_-(2t)} < 1 + \frac{4(\sqrt{2}-1)}{\sqrt{\alpha}} \left(\frac{t}{\alpha s}\right)^{1/q-1/2}.
\end{equation}
When $\alpha =1$, Eqn. (\ref{eqn:45}) is identical to the estimation condition of \citet{foucart2009sparsest}. Hence, \citet{foucart2009sparsest} can be regarded as a special case of our theory. For LSP, we have
\hint{eqn:18}
\begin{equation}\label{eqn:18}
H_r(\rho_0, \alpha , s,t) = \sqrt{\frac{s}{t}} \frac{(1+\rho_0/(\lambda \gamma))^{\frac{1}{s}}-1}{(1+\rho_0/(\lambda \gamma))^{\frac{\alpha}{t}}-1} = \sqrt{\frac{s}{t}} \frac{(\gamma\sqrt{\xi})^{-1/s} -1 }{(\gamma\sqrt{\xi})^{-\alpha/t} -1}.
\end{equation}
It should be noted that $H_r(\rho_0, \alpha , s,t) \to \infty$ as $\gamma \to 0$ for approximate MCP, $\ell_q$-norm and LSP. Figure \ref{fig:5} shows some special cases of $H_r(\rho_0,\alpha,s,t)$ for these three regularizers and $\ell_1$-norm. In Figure \ref{fig:5}, the SE conditions in Eqn. (\ref{eqn:5}) are much weaker than that of $\ell_1$-norm.

Theorem \ref{thm:1} reveals that the upper bound constraint for $\kappa_+(2t)/\kappa_-(2t)$ tends to infinity as $\gamma \to 0$ for proper non-convex regularizers. It implies that if
\hint{eqn:28}
\begin{equation}\label{eqn:28}
\kappa_-(2t) = \inf_\theta \left\{ \frac{\|X\Delta\|^2_2}{n\|\Delta\|^2_2} : \|\Delta\|_0 \le 2t \right\} >0,
\end{equation}
there exists $\gamma>0$ so that the SE condition (Eqn. (\ref{eqn:5})) is satisfied. Based on this observation, we have Corollary \ref{coro:1}. In Corollary \ref{coro:1}, $\kappa_-(2\alpha s +2) >0$ holds if the columns of $X$ are in general position\footnote{General position means any $n$ columns of $X$ are linear independent. The columns of $X$ are in general position with probability 1 if the elements of $X$ are i.i.d. drawn from some distribution, e.g., Gaussian.} and $2\alpha s +2 \le n$, which is almost optimal in the sense that it is the same as the SE condition of $\ell_0$-regularized regression \cite{zhang2011general}.
%Hence, the SE conditions of the approximate MCP, $\ell_q$-norm and LSP are much weaker than that of $\ell_1$-norm.
%Under our framework of non-convex regularizers, the regularized regression with non-convex regularizers is better than $\ell_1$-regularized regression in the sense of estimation conditions.

\subsection{Comparison between SE and RE}\label{sec:7}\hint{sec:7}

Like SE, RE is also popular to construct estimation conditions.
%RE is a regularizer-related quantity.
There are some variants of RE, e.g., $\ell_2$-RE \cite{bickel2009simultaneous, koltchinskii2009dantzig} and RIF \cite{ye2010rate, zhang2011general}.
It can derive a simple expression to the parameter estimation and the corresponding estimation condition.

\hint{def:7}
\begin{definition}[$\ell_2$-RE]\label{def:7}
For $\alpha \ge 1$, a regularizer $\mathcal R$,  an index set $\mathcal S \subset \{1, \cdots p\}$ and its complement set $\bar{\mathcal S}$, the $\ell_2$-RE is defined as
\hint{eqn:29}
\begin{equation}\label{eqn:29}
\text{RE}^{\mathcal R} (\alpha,\mathcal S) = \inf_{\Delta} \left\{ \frac{\|X\Delta\|_2^2}{n\|\Delta\|_2^2}: \mathcal R(\Delta_{\bar{\mathcal S}}) \le \alpha \mathcal R(\Delta_{\mathcal S}) \right\}.
\end{equation}
\end{definition}

\begin{definition}[Restricted Invertibility Factor]\label{def:1}\hint{def:1}
For $\tau\ge 1$, $\alpha \ge 1$, a regularizer $\mathcal R$, an index set $\mathcal S \subset \{1, \cdots p\}$, RIF is defined as
\[
\text{RIF}_{\tau}^{\mathcal R}(\alpha,\mathcal S) = \inf_{\Delta} \left\{ \frac{ |\mathcal S|^{1/\tau} \|X^T X \Delta \|_{\infty}}{ n \|\Delta\|_\tau } : \mathcal R(\Delta_{\bar{\mathcal S}}) \le \alpha \mathcal R(\Delta_{\mathcal S}) \right\}.
\]
\end{definition}

\begin{theorem}\label{thm:9}\hint{thm:9}
Suppose $\eta$-null consistency condition holds and $\alpha = (1+\eta)/(1-\eta)$. Then, $\|\hat \theta - \theta^*\|_2 \le \frac{ 2\alpha\sqrt{s}}{ \text{RE}^{\mathcal R} (\alpha, \mathcal S) } \dot r(0+)$. For any $\tau \ge 1$, $\|\hat \theta - \theta^*\|_\tau \le \frac{(1+\eta)\lambda^* s^{1/\tau} }{\text{RIF}^{\mathcal R}_\tau (\alpha, \mathcal S)}$.
\end{theorem}

The estimation conditions based on RE require that $\text{RE}^{\mathcal R} (\alpha, \mathcal S) > 0$ or $ \text{RIF}^{\mathcal R}_\tau (\alpha, \mathcal S) > 0$. The same conclusion also can be obtained for $\ell_1$-regularized regression \cite{negahban2012unified, zhang2011general}. What we are interested in is whether non-convex regularizers allow a larger value of $\text{RE}^{\mathcal R}(\alpha, \mathcal S)$ than $\ell_1$-norm, i.e., whether $\text{RE}^{\mathcal R}(\alpha, \mathcal S)>0$ becomes weaker by employing non-convex regularizers.

Define $\Omega(\beta)=\{\Delta \in \mathbb R^n: \mathcal R(\beta \Delta_{\bar{\mathcal S}}) \le \alpha \mathcal R(\beta \Delta_{\mathcal S}), \|\Delta\|_2=1\} $ for $\beta>0 $. The concavity of $r(u)$ gives that $\dot r(0+)u \ge r(u) \ge u \dot r(u-)$, which derives that
\[
\Omega(\beta) \supset
 \{\Delta \in \mathbb R^n:
\dot r(0+) \beta \|\Delta_{\bar{\mathcal S}}\|_1 \le
\alpha \beta \langle |\Delta_{\mathcal S}|, \dot r(|\beta \Delta_{\mathcal S}|-) \rangle , \|\Delta\|_2=1 \} ,
\]
where $|\Delta_{\mathcal S}|$ is the vector composed of the absolute values of the components of  $\Delta_{\mathcal S}$, i.e., $|\Delta_{\mathcal S}| = (|\Delta_i|: i\in \mathcal S)$. In the same way, $\dot r(|\beta \Delta_{\mathcal S}|-) = (\dot r(|\beta \Delta_i|-): i\in \mathcal S)$. Thus, we give an upper bound to $\text{RE}^{\mathcal R} (\alpha,\mathcal S)$:
\begin{align}
\text{RE}^{\mathcal R} (\alpha,\mathcal S) & = \inf_{\beta > 0, \Delta \in \mathbb R^ p}\{ \frac{\|X \Delta\|_2^2}{n \|\Delta\|_2^2}: \Delta \in \Omega(\beta)\} \nonumber\\
& \le \inf_{\beta > 0, \Delta \in \mathbb R^ p} \{ \frac{\|X \Delta\|_2^2}{n \|\Delta\|_2^2} : \dot r(0+) \|\Delta_{\bar{\mathcal S}}\|_1 \le
\alpha \langle |\Delta_{\mathcal S}|, \dot r(|\beta \Delta_{\mathcal S}|-) \rangle, \|\Delta\|_2=1  \} \nonumber\\
& \stackrel{(\beta\to 0+)}{\le} \inf_{\Delta \in \mathbb R^ p} \{ \frac{\|X \Delta\|_2^2}{n \|\Delta\|_2^2}: \|\Delta_{\bar{\mathcal S}}\|_1 \le
\alpha \|\Delta_{\mathcal S}\|_1, \|\Delta\|_2=1 \} \nonumber\\
%& = \inf_\Delta \{  \frac{\|X \Delta\|_2^2}{n \|\Delta\|_2^2} : \| \Delta_{\bar{\mathcal S}} \|_1 \le \alpha \| \Delta_{\mathcal S} \|_1\} \nonumber\\
& = \text{RE}^{\ell_1} (\alpha,\mathcal S) \nonumber
\end{align}

$\text{RE}^{\mathcal R} (\alpha,\mathcal S) \le \text{RE}^{\ell_1} (\alpha,\mathcal S)$ means that the RE based condition of non-convex regularized regression $\text{RE}^{\mathcal R} (\alpha,\mathcal S)>0$ is not relaxed. \citet{negahban2012unified} put an additional constraint $\mathcal U(\epsilon) = \{\Delta: \|\Delta\| \ge \epsilon\}$ to the definition of RE. This constraint avoids the bad case $\Delta \to 0$. However, it still cannot guarantee to provide larger RE for non-convex regularizers than $\ell_1$-norm. For example, let $t_1, t_2$ and $t_3$ satisfy that $|t_1| + |t_2| \le 2|t_3|$ and $\alpha=2$, $\mathcal S=\{3\}$ and $\bar{\mathcal S}=\{1,2\}$. Thus, the concavity of $r(u)$ implies that $r(|t_1|) + r(|t_2|) \le 2r((|t_1| + |t_2|)/2) \le 2r(|t_3|)$. For this case, $\{\Delta: \|\Delta_{\bar{\mathcal S}}\|_1 \le \alpha \|\Delta_{\mathcal S}\|_1\} \subset \{\Delta: \mathcal R (\Delta_{\bar{\mathcal S}}) \le \alpha \mathcal R (\Delta_{\mathcal S}) \}$. Thus, $\text{RE}^{\mathcal R} (\alpha, \mathcal S) \le \text{RE}^{\ell_1} (\alpha, \mathcal S)$. For RIF, we have the same result.
Although non-convex regularizers give better approximations to $\ell_0$-norm, the RE of non-convex regularizers cannot be guaranteed to be lager than that of $\ell_1$-norm. The framework of RE does not leave space to relax the estimation condition for non-convex regularizers.

The only difference between the definitions of SE and RE lies in the constraints for $\Delta$. The two constraints $\{\Delta: \|\Delta\|_0 \le 2t\}$ and $\{\Delta: \mathcal R(\Delta_{\bar{\mathcal S}}) \le \alpha \mathcal R(\Delta_{\mathcal S})$ do not contain each other. However, we observe that $\kappa_-(2t) \ge \min_{|\mathcal T|\le s} \text{RE}^{\mathcal R} ((2t-s)/s ,\mathcal T) \ge \min_{|\mathcal T|\le s} \text{RE}^{\mathcal R} (2\alpha-1+2/s,\mathcal T)$ for $t \ge \alpha s+1$. When $\eta$ is small and $s\gg 2$, $2\alpha -1 +2/s$ is close to $\alpha$ and $\min_{|\mathcal T|\le s} \text{RE}^{\mathcal R} (2\alpha-1+2/s,\mathcal T) \approx \min_{|\mathcal T|\le s} \text{RE}^{\mathcal R} (\alpha,\mathcal T)$. Hence, with proper regularizers, the SE condition in Eqn. (\ref{eqn:28}) is a weaker condition than $\min_{|\mathcal T| \le s} \text{RE}^{\mathcal R} (\alpha ,\mathcal T)>0$.

We can also compare RE and our SE conditions with the help of the failure bound of RIC $\delta_{2s} = 1/\sqrt{2}$ for $\ell_1$-minimization recovery \cite{Davies2009Restricted}, where $\ell_1$-minimization recovery includes the basis pursuit \cite{chen1999atomic} and Dantzig selector \cite{candes2007dantzig}. The failure bound means that for any $\varepsilon>0$ there exists $X \in \mathbb R^{(p-1)\times p}$ with $\delta_{2s}<1/\sqrt{2} + \varepsilon$ where $\ell_1$-minimization recovery fails. On the other hand, $\ell_1$-minimization recovery succeeds when $\text{RE}^{\ell_1}(\alpha,\mathcal S) >0$ \cite{bickel2009simultaneous}, like $\ell_1$-regularized regression (Theorem \ref{thm:9}). Thus, $\min_{|\mathcal T| \le s} \text{RE}^{\ell_1}(\alpha,\mathcal T) = 0$ if $\delta_{2s} \ge 1/\sqrt{2}$, i.e., $\kappa_+(2s)/\kappa_-(2s) \ge 3+ 2\sqrt{2}$. Since non-convex regularizers cannot weaken RE conditions, $\kappa_+(2s)/\kappa_-(2s) \ge 3+ 2\sqrt{2}$ also causes $\min_{|\mathcal T| \le s} \text{RE}^{\mathcal R}(\alpha,\mathcal T) = 0$ for non-convex regularizers. On the contrary, our SE conditions, e.g., $\kappa_-(2\alpha s+2)>0$, still hold with proper non-convex regularizers even when $\kappa_+(2s)/\kappa_-(2s) \ge 3+ 2\sqrt{2}$.

\subsection{Comparison with the conditions for feature selection}

\citet{shen2013constrained} gave a necessary condition for consistent feature selection, which can be relaxed further to $\kappa_-(s) > C\log p/n$ with a constant $C>0$ independent of $p,~ s,~ n$. This necessary condition needs $\kappa_+(s) / \kappa_-(s)$ to be upper bounded by a constant which is independent of the regularizers. For their DC algorithm based methods, they tightened the conditions to that $\kappa_+(2\tilde s)/\kappa_-(2\tilde s)$ is upper bounded, where $\tilde s$ is the number of non-zero components of the solutions given by their methods. This condition cannot be verified until the solutions are given.
%or the same RE conditions as LASSO are given \cite{zhang2012general}.
However, our SE conditions do not depend on the sparseness of the practical solutions (see Section \ref{sec:6}).

\section{Sparse Estimation of AGAS Solutions}\label{sec:6}\hint{sec:6}

For Problem (\ref{eqn:1}), it is practical to obtain a solution which is approximate global (AG) (Definition \ref{def:2}) and approximate stationary (AS) (Definition \ref{def:3}). We show in this section that this kind of solutions also give good estimation to the true parameters.

\begin{definition}\label{def:2}\hint{def:2}
Given $\mu\ge 0$, we say $\tilde \theta$ is a $(\theta^*, \mu)$-approximate global solution of $\min_\theta \mathcal F (\theta)$ if $\mathcal F (\tilde \theta) \le \mathcal F (\theta^*) + \mu$.
\end{definition}

\begin{definition}\label{def:3}\hint{def:3}
Given $\nu\ge 0$, we say $\tilde \theta$ is a $\nu$-approximate stationary solution of $\min_\theta \mathcal F (\theta)$ if the directional derivative of $\mathcal F$ at $\tilde \theta$ in any direction $d \in \mathbb R^{p}$ with $\|d\|_2=1$ is no less than $-\nu$, i.e., $\mathcal F'(\theta; d) \ge -\nu$.
\end{definition}

The directional derivative is defined as $\mathcal F'(\theta;d) = \liminf_{\lambda \downarrow 0} (\mathcal F(\theta+\lambda d) - \mathcal F(\theta) )/\lambda$ for any $\theta \in \mathbb R^p$ and $d\in \mathbb R^p$. For Problem (\ref{eqn:1}), $\mathcal F'(\theta;d) =d^T \nabla \mathcal L(\theta) + \sum_{i=1}^p \mathcal R'(\theta_i;d_i)$.

The following theorem gives the parameter estimation result with AGAS solutions. Let  $\tilde u_0 \ge 0$ be the zero gap of $\tilde \theta$ and $\tilde \rho_0 =\min\{ \tilde u_0, \min_{i \in \text{supp}(\theta^*)} |\theta^*_i|\} $.

\begin{theorem}[Parameter Estimation of AGAS solutions]\label{thm:3}
\hint{thm:3}
Suppose the following conditions hold for the regularized regression.
\begin{enumerate}
\item $\tilde \theta$ is a $(\theta^*, \mu)$-AG solution and $\nu$-AS solution.
\item $r(u)$ is invertible for $u \ge 0$ and $r^{-1}(u/s_1) / r^{-1}(u/s_2)$ is a non-decreasing function w.r.t. $u$ for any $s_2 \ge s_1 >0$;
\item The regularized regression satisfies $\eta$-null consistency;
\item The following SE condition holds for some integer $t \ge \alpha s+1$,
%\hint{eqn:6}
    \begin{equation}\label{eqn:6}
    \kappa_+(2t) / \kappa_-(2t) < 4(\sqrt{2}-1) G_r(\tilde\rho_0,\alpha, s,t)+ 1,
    \end{equation}
    where $\alpha = \frac{1+\eta}{ 1- \eta}$, $G_r(\tilde\rho_0,\alpha, s,t) =  \frac{\sqrt{st}}{t-1} \frac{r^{-1}(r(\tilde\rho_0)/ s)}{r^{-1}( \alpha r(\tilde\rho_0)/(t-1))}$ for $\tilde\rho_0 >0$ and $G_r(0,$ $\alpha, s,t) =  \lim_{\rho \to 0+} G_r(\rho, \alpha ,s,t)$.
\end{enumerate}
Then, $\|\tilde \theta - \theta^*\|_2 \le C_4 \tilde\epsilon + C_5 r^{-1}(\frac{\mu}{1-\eta})$, where $\tilde\epsilon=\dot r(0+) +  \eta \lambda^* + \nu$ and $C_4,~C_5$ are positive constants. $C_4$ and $C_5$ are defined in Eqn. (\ref{eqn:41}) and (\ref{eqn:42}).
\end{theorem}

The condition 2, 3 and 4 are almost the same as the three conditions of Theorem \ref{thm:1} except the slightly different requirements for $t$ and the definition of $G_r(\tilde\rho_0, \alpha, s, t)$. Consequently, the discussion in Section \ref{sec:12} is also suitable for this theorem:
\begin{enumerate}
\item The non-invertible basis functions can be approximated by approximate invertible basis functions;
\item Without $\xi$-sharp concavity, the condition 4 of Theorem \ref{thm:3} is almost the same as RIP conditions in \citet{foucart2009sparsest};
\item With $\xi$-sharp concavity and a positive zero gap (we show in Theorem \ref{thm:5} that our CD methods guarantee the positive zero gaps), SE based estimation conditions can be much relaxed.
\end{enumerate}

Theorem \ref{thm:3} shows that the error bounds of parameter estimation are mainly determined by four parts: the slope of $r(u)$ at zero $\dot r(0+)$, the parameter $\lambda^* = O(\epsilon/\sqrt{n})$, the degree of approximating the stationary solutions $\nu$ and the degree of approximating the global optimums $r^{-1}(\mu/(1-\eta))$. If $r(u)=\lambda^2 r_0(u/\lambda;\gamma)$ and $r_0(u; \gamma)$ has a finite derivative at zero, we know that $\dot r(0+) = \lambda \dot r_0(0+;\gamma)$, e.g., $\dot r(0+) = \lambda$ for MCP. Since $\lambda = O(\epsilon/\sqrt{n})$ by Eqn. (\ref{eqn:20}) in this paper, the estimation error bound is actually
\[
\|\tilde \theta - \theta^* \|_2 \le O(\epsilon/\sqrt{n}) + O(\nu) + O(r^{-1}(\mu/(1-\eta))).
\]
According to Theorem \ref{thm:3}, we do not need to solve Problem (\ref{eqn:1}) exactly. A good suboptimal solution is enough to give good parameter estimation. Even, we do not need a strictly stationary solution since Theorem \ref{thm:3} allows a margin $\nu$. So, the non-convex regularized regression is robust to the inaccuracy of the solutions, which is important for numerical computation.

It should be noted that $\dot r(0+)$ is required to be finite in Theorem \ref{thm:3}, which forbids the regularizers with infinite $\dot r(0+)$, e.g., $\ell_0$-norm and $\ell_q$-norm ($0<q<1$). It may be due to the strongly NP-hard property brought by $\ell_0$-norm and $\ell_q$-norm regularized regression \cite{chen2011complexity}.

Similar to Theorem \ref{thm:8}, we give the following sparseness estimation result for AGAS solutions. The proof is the same as that of Theorem \ref{thm:8}.

\begin{theorem}[Sparseness Estimation of AGAS solutions]\label{thm:2}
%\hint{thm:2}
Suppose the conditions of Theorem \ref{thm:3} hold. Let $b=(t-1) r \left( c_4 \tilde\epsilon + c_5 r^{-1}\left( \frac{\mu}{1-\eta}\right)\right)$,
where $c_4$ and $c_5$ are defined in Eqn. (\ref{eqn:43}) and Eqn. (\ref{eqn:44}). Consider $l_0>0$ and integer $m_0>0$ such that
\[
\sqrt{ \frac{2\kappa_+(m_0)}{m_0} (\frac{\mu}{1-\eta} + b) } + \|X^T e/n\|_\infty \le \dot r(l_0-).
\]
Then, $|\text{supp}(\tilde \theta) \backslash \mathcal S| <  m_0 + b/r(l_0)$.
\end{theorem}

The sparseness of AGAS solutions is also affected by $\tilde\epsilon=\dot r(0+) + \eta \lambda^* + \nu$ and $\mu$. Theorem \ref{thm:2} can also derive a similar conclusion as Corollary \ref{coro:2}. For an AGAS solution with small $\nu$ and $\mu$, the sparseness of the solution is on the order of $s$, just like the global solutions.

\subsection{Approximate Global Solutions} \label{sec:3}
%\hint{sec:3}

We need AG solutions in Theorem \ref{thm:3} and Theorem \ref{thm:2}. The methods to obtain such solutions are crucial consequently.
%With the conditions of $\eta$-null consistency and $\text{RIF}_1^{\ell_1} ((1+\eta)/(1-\eta) , \mathcal S) > 0$, the solution of LASSO $\hat\theta^{\ell_1}$ with $\lambda  = \lambda^*$ can be an $O(\lambda^2 s)$-AG solution if $\|\hat\theta^{\ell_1}\|_0$ is of the order of $s$ \cite{zhang2011general}. However, the conditions are so strong that LASSO can yield consistent parameter estimations by itself. When the RE conditions fail, it is difficult to theoretically guarantee the quality of the solutions given by LASSO.
Instead of restricting to the solutions given by a specific algorithm, we use the prediction error $\|X\theta^0 - y\|_2^2/(2n)$ to give a quality guarantee for any solution $\theta^0$ that is regarded as an AG solution.

\begin{theorem}\label{thm:11}\hint{thm:11}
Suppose $\theta^0$ is an $s_0$-sparse vector with the prediction error $\mu_0^2 = \|X \theta^0 - y\|_2^2/(2n)$. If $\kappa_-(s+s_0)>0$, then $\theta^0$ is a ($\theta^*$, $\mu$)-AG solution where
\[
\mu =\mu_0^2 + (s+s_0) r \left( \frac{\sqrt{2} \mu_0 + \epsilon/\sqrt{n}}{\sqrt{(s+s_0) \kappa_{-}(s+s_0)}} \right)
\]
\end{theorem}

\begin{corollary}\label{coro:3}
\hint{coro:3}
Suppose $\theta^0$ is an $s_0$-sparse vector with the prediction error $\mu_0 = \zeta \epsilon/\sqrt{n}$ for some $\zeta \ge 0$ and the basis function has the formulation $r(u) = \lambda^2 r_0(u/\lambda)$ with $\lambda=\eta^{-1} b_0 \epsilon / \sqrt{n}$. Then, $\theta^0$ is a $(\theta^*, C_6 \epsilon^2/n)$-AG solution where
$$C_6 = \zeta^2  +  \frac{(s+s_0) b_0^2}{\eta^2} r_0 \left( \frac{(1+ \sqrt{2} \zeta)\eta }{ b_0 \sqrt{(s+s_0) \kappa_{-}(s+s_0)}} \right) .$$
\end{corollary}

The methods that explicitly control the sparseness of its solutions are suitable for giving the AG solutions, e.g., OMP \cite{tropp2007signal} and GraDeS \cite{garg2009gradient}. However, we do not need the strong conditions for consistent parameter estimation for these methods, e.g., $\delta_{2s}<1/3$ for GraDeS \cite{garg2009gradient} or $ (\kappa_+(1)/\kappa_-(t)) \log (\kappa_+(s)/\kappa_-(t))$ grows sub-linearly as $t$ for OMP \cite{zhang2011sparse}. In fact, Theorem \ref{thm:11} only requires $\kappa_{-}(s+s_0)>0$. Hence, $s_0$ can be large enough to make $\mu_0$ to be small. The relationship between $\mu_0$ and $s_0$ depends on the employed method and the design matrix $X$. Even with a bad value of $\mu$ in the initialization, we can decrease it further by CD methods as stated in Section \ref{sec:4}.

\subsection{Approximate Stationary Solutions with Zero Gap} \label{sec:4}\hint{sec:4}

Theorem \ref{thm:3} also requires the solution to be $\nu$-AS and has a positive zero gap. General gradient descent algorithms can provide stationary solutions but they cannot ensure a positive zero gap. However, we observe that the coordinate descent (CD) methods can yield AS solutions and all of these solutions have positive zero gaps under proper sharp concavity conditions.

In every step, CD only optimizes for one dimension, i.e.,
%\hint{eqn:22}
\begin{equation}\label{eqn:22}
\theta^{(k)}_i = \arg\min_{u \in \mathbb R} \mathcal F((\theta^{(k)}_1, \cdots, \theta^{(k)}_{i-1}, u, \theta^{(k-1)}_{i+1}, \cdots, \theta^{(k-1)}_{p} )^T ) + \frac{\psi}{2}(u - \theta_i^{(k-1)})^2,
\end{equation}
where $k$ is the number of iterations, $i=1, \cdots,p$ and $\psi>0$ is a positive constant. The constant $\psi$ plays a role of balance between decreasing $\mathcal F(\theta)$ and not going far from the previous step. The above CD method is also called \emph{proximal coordinate descent}. For Problem (\ref{eqn:1}), the CD methods iterate as follows.
\hint{eqn:23}
\begin{equation}\label{eqn:23}
\theta^{(k)}_i = \arg\min_{u \in \mathbb R} \frac{1}{2} \left( \frac{\|x_i\|_2^2}{n} +  \psi \right)
\left( u  - \frac{\psi \theta_i^{(k-1)} + x_i^T \omega_i^{(k)}/n}{\psi + \|x_i\|_2^2/n } \right)^2 +  r(|u|),
\end{equation}
where $x_i$ is the i-th column of the design matrix $X$ and $\omega_i^{(k)} = y - \sum_{j<i} x_j \theta^{(k)}_j - \sum_{j>i} x_j \theta^{(k-1)}_j$. Problem (\ref{eqn:23}) is a non-convex but only one-dimensional problem. All of its solutions are between $0$ and $\frac{\psi \theta_i^{(k-1)} + x_i^T \omega_i^{(k)}/n}{\psi + \|x_i\|_2^2/n }$. We assume that Problem (\ref{eqn:23}) can be exactly solved. If Problem (\ref{eqn:23}) has more than one minimizer, any one of them can be selected as $\theta_i^{(k)}$. In this paper, CD methods stop iterating if
\hint{eqn:55}
\begin{equation}\label{eqn:55}
\|\theta^{(k)} - \theta^{(k-1)}\|_2 \le \tau,
\end{equation}
where $\tau>0$ is a small tolerance proportional to the value $\nu$ (see Theorem \ref{thm:6}).

\begin{theorem}\label{thm:5}
\hint{thm:5}
If $r(u)$ is $(\xi+\psi)$-sharp concave over $(0,u_0)$, then $\theta^{(k)}_i \ge u_0$  or $\theta^{(k)}_i = 0$ for any $k=1,2,\cdots$ and any $i=1,\cdots, p$.
\end{theorem}

The above zero gap property of CD is a corollary of Theorem \ref{thm:4}. The sharp concavity condition of Theorem \ref{thm:5} is a little stronger than the requirements of Theorem \ref{thm:1}. Nonetheless, we can set $\psi$ to be small to narrow the difference between the sharp concavity conditions of Theorem \ref{thm:1} and Theorem \ref{thm:5}.

Besides the zero gap, we show in the following theorem that CD methods simultaneously give AS solutions and keep them to be still AG solutions.

\begin{theorem}\label{thm:6}
\hint{thm:6}
$\{\mathcal F(\theta^{(k)})\}$ is a non-increasing sequence and converges;
For any $\nu>0$ and with $\tau = \nu/(\sqrt{p}(\psi + p \xi))$, CD stops within $k=1 + \frac{2p (\psi + p \xi)^2 \mathcal F(\theta^{(0)})}{\psi \nu^2}$ iterations and outputs a $\nu$-AS solution, where $p$ is the number of columns of the design matrix $X$.
\end{theorem}

Theorem \ref{thm:6} shows CD methods give a further decrease to the value $\mu$ of AG property and guarantees the $\nu$-AS property, which is necessary for sparse estimation in Theorem \ref{thm:3} and Theorem \ref{thm:2}. This theorem also gives an upper bound for $\nu$, i.e.,
\hint{eqn:56}
\begin{equation}\label{eqn:56}
\nu \le \sqrt{p}(\psi + p \xi)\|\theta^{(k)} - \theta^{(k-1)}\|_2,
\end{equation}
where $k$ is the number of iterations. Usually, we hope $\nu$ is on the order of $\lambda^*$ so that $\tilde\epsilon = \dot r(0+) + \eta \lambda^* + \nu = O(\lambda^*) = O(\epsilon/\sqrt{n})$ in Theorem \ref{thm:3}.

CD has been applied to the non-convex regularized regression by \citet{breheny2011coordinate} and \citet{mazumder2011sparsenet} . However, their non-convex regularizers are restrictive because they need Eqn. (\ref{eqn:23}) to be strictly convex for $\psi=0$. They could not deal with the MCP with $\gamma \le 1$, the SCAD with $\gamma \le 2$ or the LSP with $\gamma \le 1$. Compared with them, the conclusions of Theorem \ref{thm:6} are weaker but they are enough to obtain $\nu$-AS solutions and the regularizers can approximate $\ell_0$-norm arbitrarily.

%Finally, we are interested in whether the parameter estimation given by the AGAS solutions is optimal as the global solutions. Suppose the solution is $\nu$-AS with $\nu= O(\epsilon/\sqrt{n})$ and $(\theta^*,\mu)$-AG with $\mu = C_6 \epsilon^2/(2n)$ as stated in Corollary \ref{coro:3}. In fact, $\mu$ should be less than $C_6 \epsilon^2/(2n)$ since CD can further give it a decrease. Let the noise be white noise with the variance $\sigma$ and $n = O(s \log p)$ so that the SE condition in Eqn. (\ref{eqn:6}) holds for Gaussian design matrix even with $\tilde\rho_0=0$. For $r(u) = \lambda^2 r_0(u/\lambda)$ and $\lambda = b_0 \epsilon/\sqrt{n}$, we have $\|\tilde \theta - \theta^*\|_2 \le O(\sigma s \sqrt{\log p/n})$ according to Theorem \ref{thm:3}, which means the AGAS solutions are also almost optimal.
%If $\tilde\rho_0 \ge O(\lambda)$, the oracle property can be induced with appropriate additional conditions \cite{zhang2011general}.

\section{Experiment}\label{sec:11}
\hint{sec:11}

In this section, we experimentally show the performance of CD methods on giving AGAS solutions and the degree of weakness of the estimation conditions required by the sharp concave regularizers.

\subsection{AGAS solutions}\label{sec:10}
\hint{sec:10}

In Section \ref{sec:6}, we prove that $\mu$ is monotonously decreasing, $\nu$ tends to 0 and the zero gap $\tilde u_0$ is maintained in each iteration of CD algorithm. We experimentally show these in this part.

We set the dimension of the parameter as $p=1000$, the number of non-zero components of $\theta^*$ (the true parameter) $s=\log p$. We randomly choose $s$ indices as the non-zero components. The non-zero components are i.i.d. drawn from $\mathcal N(0,1)$ and those belonging to $(-0.1,0.1)$ are promoted to $\pm 0.1$ according to their signs.

The elements of the design matrix $X \in \mathbb R^{n\times p}$ are i.i.d. drawn from $\mathcal N(0,1)$, where $n=10s\log p$.  The noise $e$ is drawn from $\mathcal N(0,I_n)$ and is normalized such that $\epsilon=\|e\|_2=0.01$. We fix $\gamma=0.1$ and $\eta=0.01$ for all the non-convex regularizers (LSP, MCP and GP) and use Eqn. (\ref{eqn:20}) to choose $\lambda$.

For CD algorithm, we set $\psi=0.1$. The CD algorithm is initialized with zero vectors and terminated when $\nu$ is below $10^{-3}$ (we set $\tau = 10^{-3}/(\sqrt{p} (\psi + p \xi))$ by Theorem \ref{thm:6}) or the number of iterations is over 500. For each regularizer, we run CD for 100 trials with independent true parameters and design matrices.

\begin{figure}
  \centering
  \subfigure[LSP]
  {
  \includegraphics[width=0.31\textwidth]{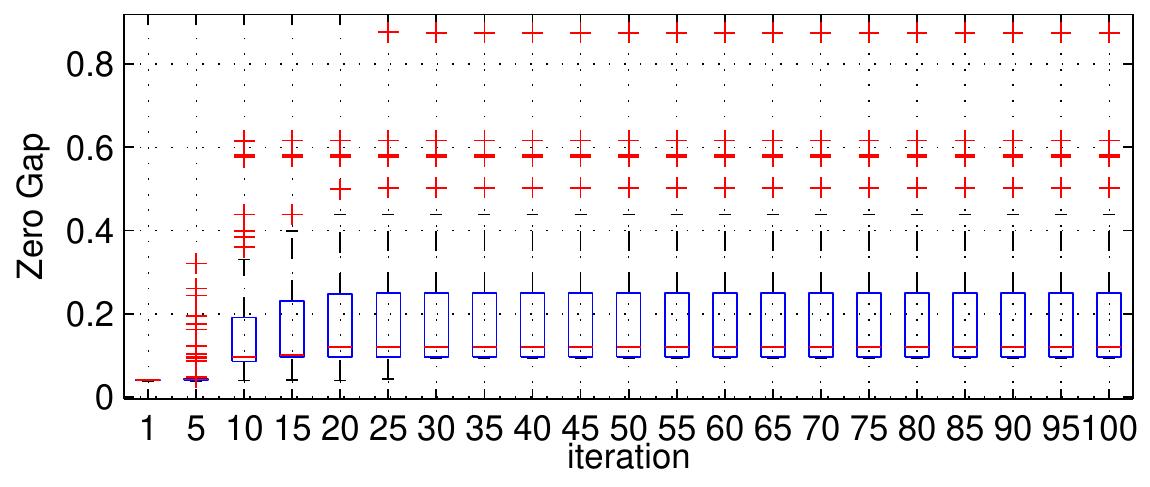}
  \includegraphics[width=0.31\textwidth]{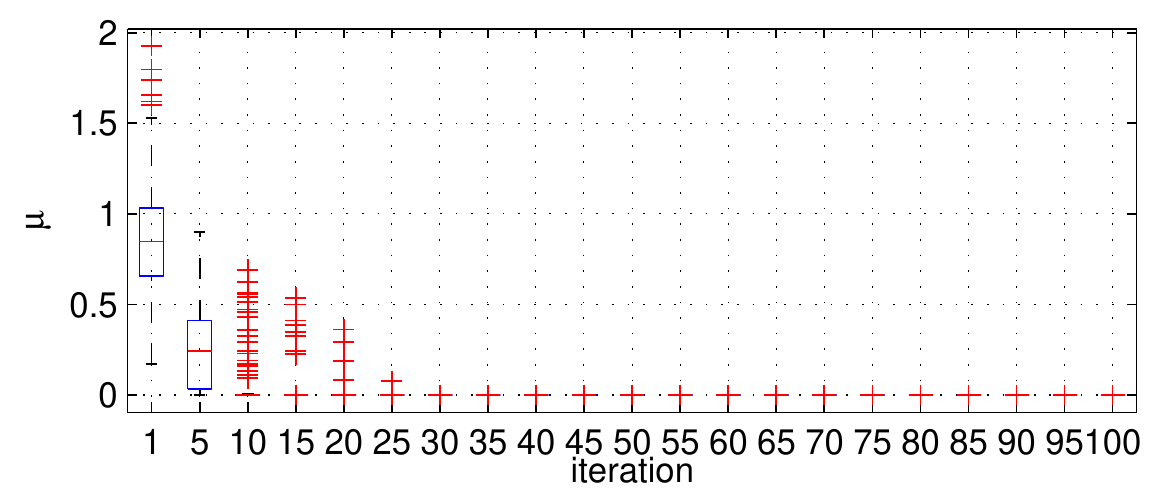}
  \includegraphics[width=0.31\textwidth]{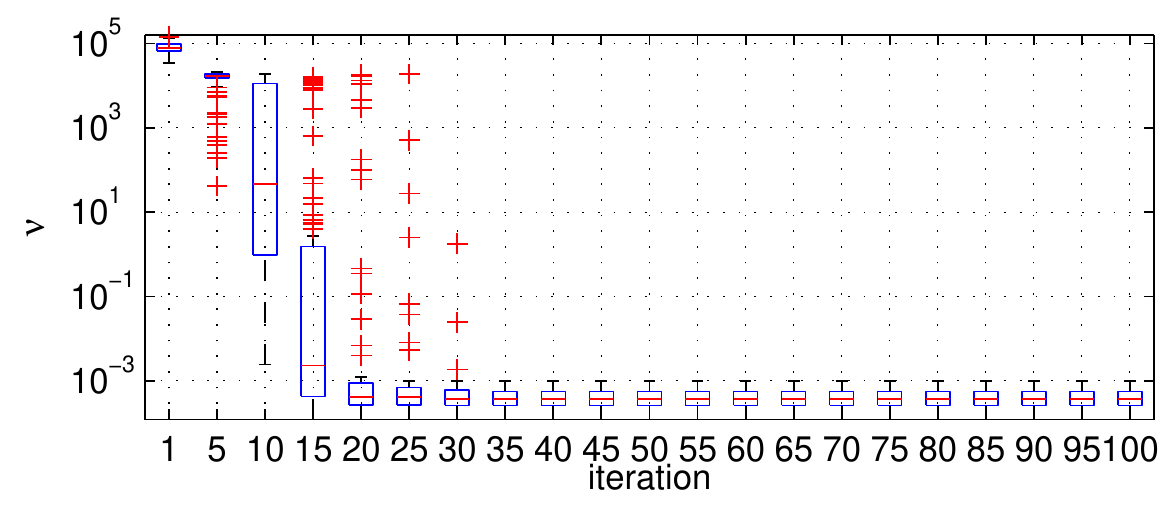}
  }

  \subfigure[MCP]
  {
  \includegraphics[width=0.31\textwidth]{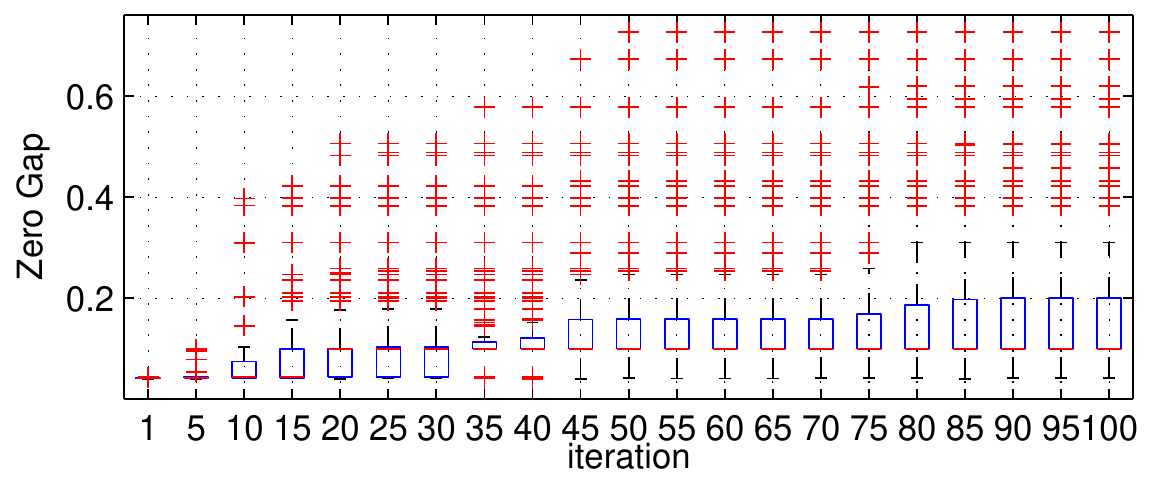}
  \includegraphics[width=0.31\textwidth]{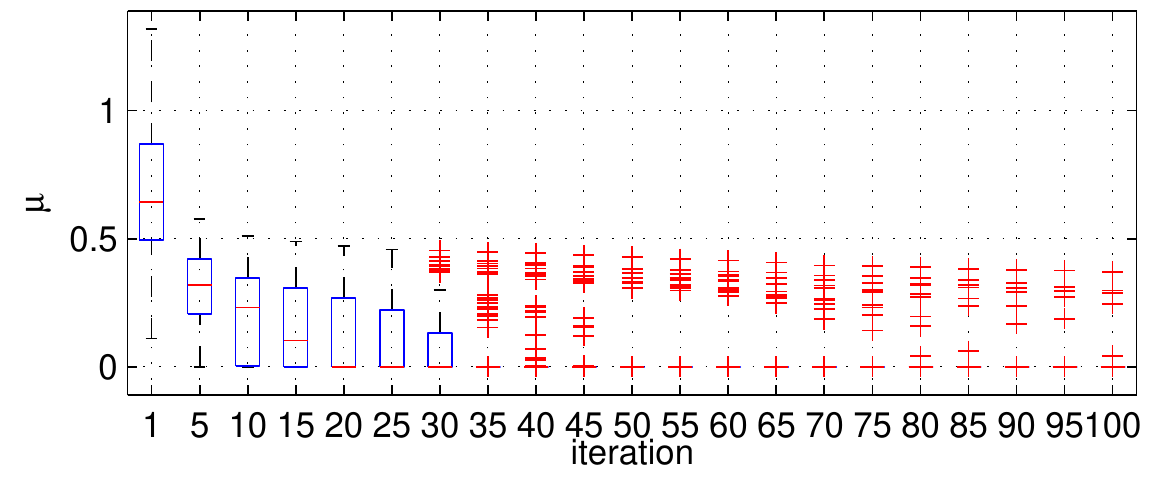}
  \includegraphics[width=0.31\textwidth]{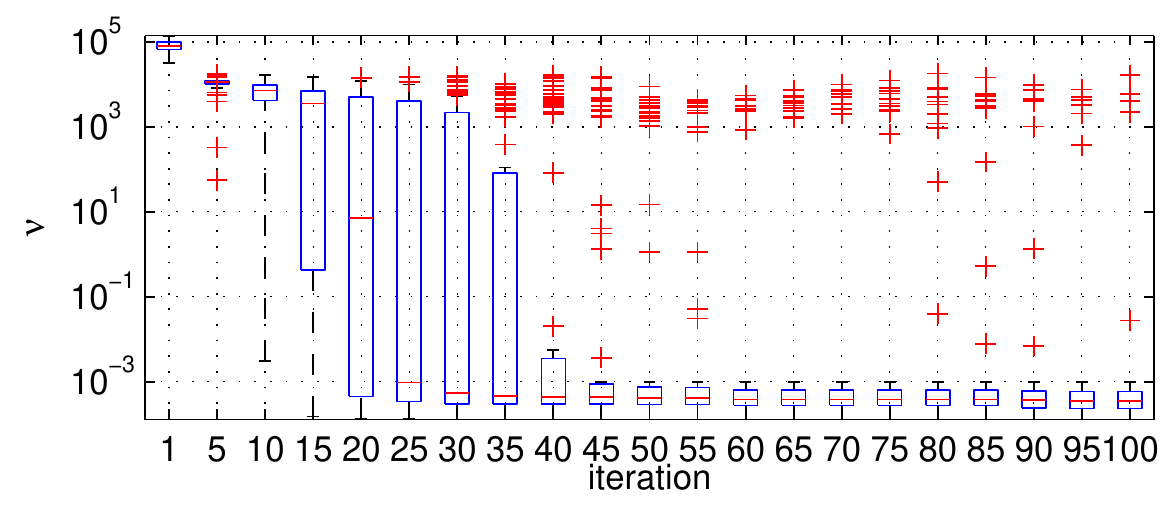}
  }

  \subfigure[GP]
  {
  \includegraphics[width=0.31\textwidth]{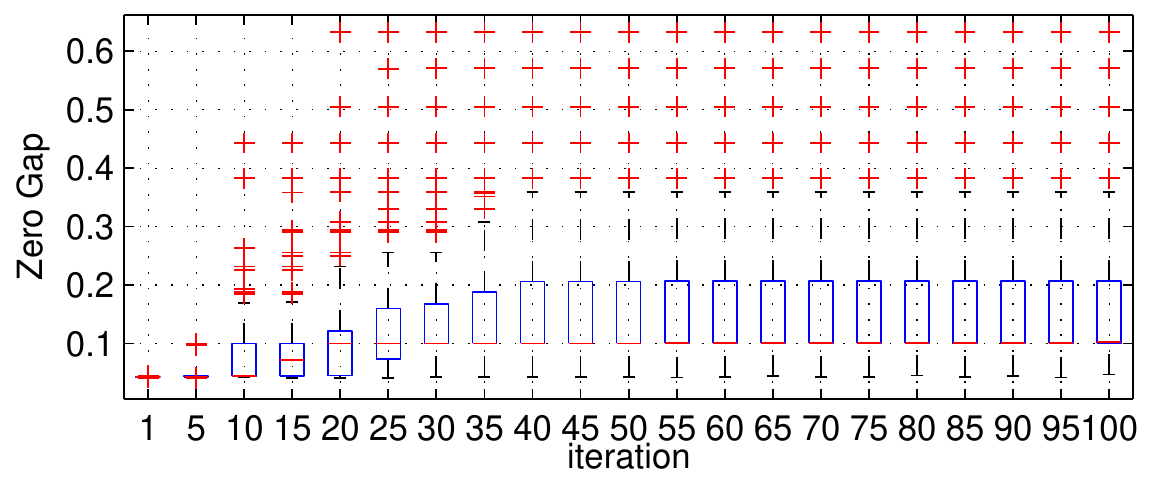}
  \includegraphics[width=0.31\textwidth]{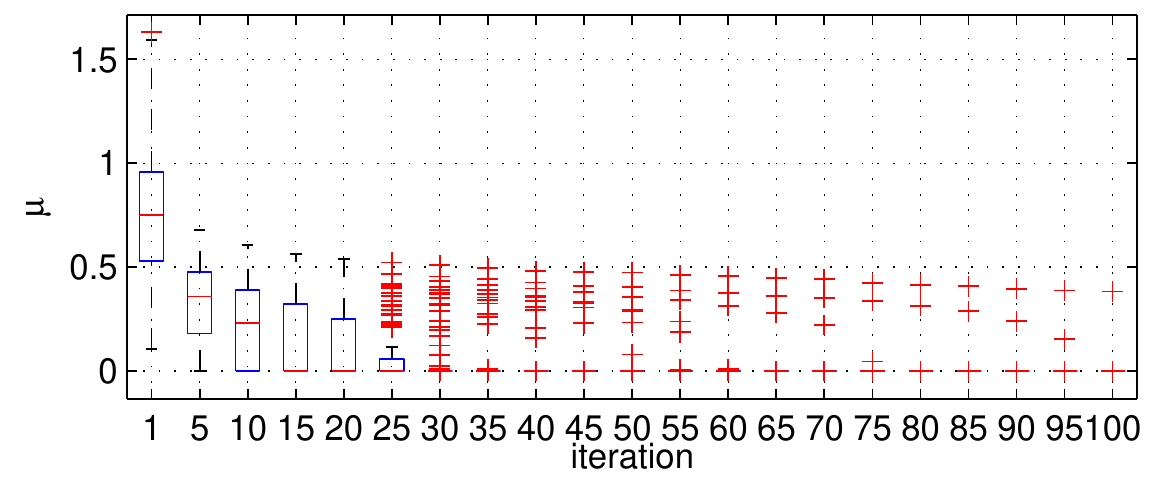}
  \includegraphics[width=0.31\textwidth]{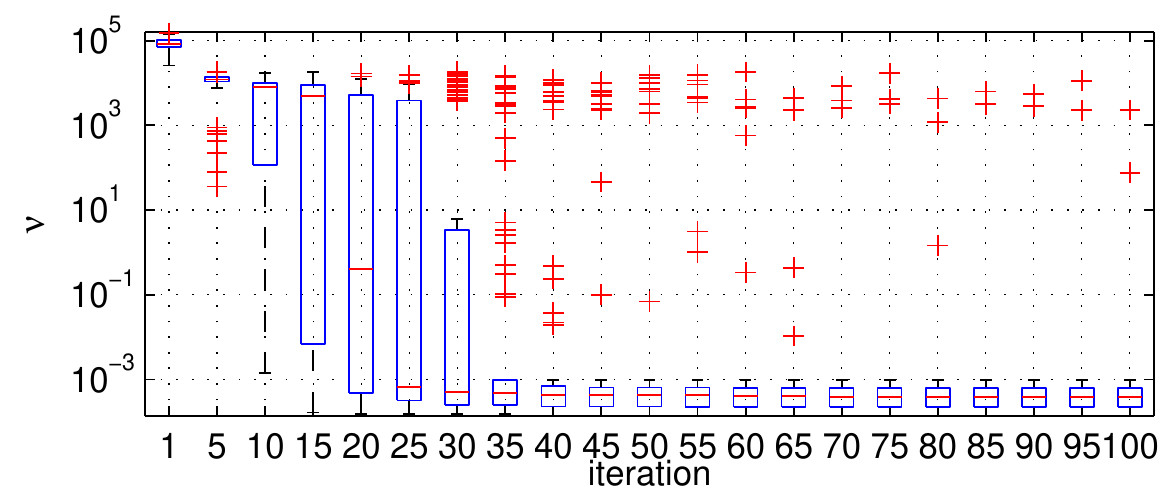}
  }
  \caption{\hint{fig:1}The zero gap $\tilde u_0$ (left) and the parameters of AGAS solutions $\mu$ (middle) and $\nu$ (right) in each iteration of CD algorithms. The figures are in the form of the boxplots of the 100 trials of CD algorithms. The right column is actually the boxplots of the upper bound for $\nu$ in Eqn. (\ref{eqn:56}).}\label{fig:1}
\end{figure}

We illustrate the boxplots for $\tilde u_0$, $\mu$ and $\nu$ of each iteration in Figure \ref{fig:1}. The left column shows that CD methods maintain the zero gaps in each iteration as stated in Theorem \ref{thm:5}.
%The minimal zero gaps after some iterations are close to 0.1 that is the minimal zero gap of the true parameters.
The middle column shows $\mathcal F(\theta^{(k)}) - \mathcal F(\theta^*)$ decrease to zero for most of trials in 100 iterations. The right column shows that most of the solutions are very close to stationary solutions within 100 iterations.

\subsection{Weaker Conditions for Sparse Estimation}

We show the performance of non-convex regularizers for sparse estimation in this part. For an estimation $\tilde\theta$, three criterions are used to describe the performance of sparse estimation: 1. sparseness $\|\tilde\theta\|_0$; 2. Relative recovery error (RRE) $\|\tilde\theta - \theta^*\|_2/\|\theta^*\|_2$; 3. Support recovery rate (SRR) $| \text{supp}(\tilde\theta) \cap \text{supp}(\theta^*)| / |\text{supp}(\tilde\theta) \cup \text{supp}(\theta^*)|$. A weaker estimation condition than convex regularizers can be verified by achieving a more accurate sparseness, lower RRE or higher SRR with less sampling size.

\begin{figure}
  \centering
  \includegraphics[width=0.32\textwidth]{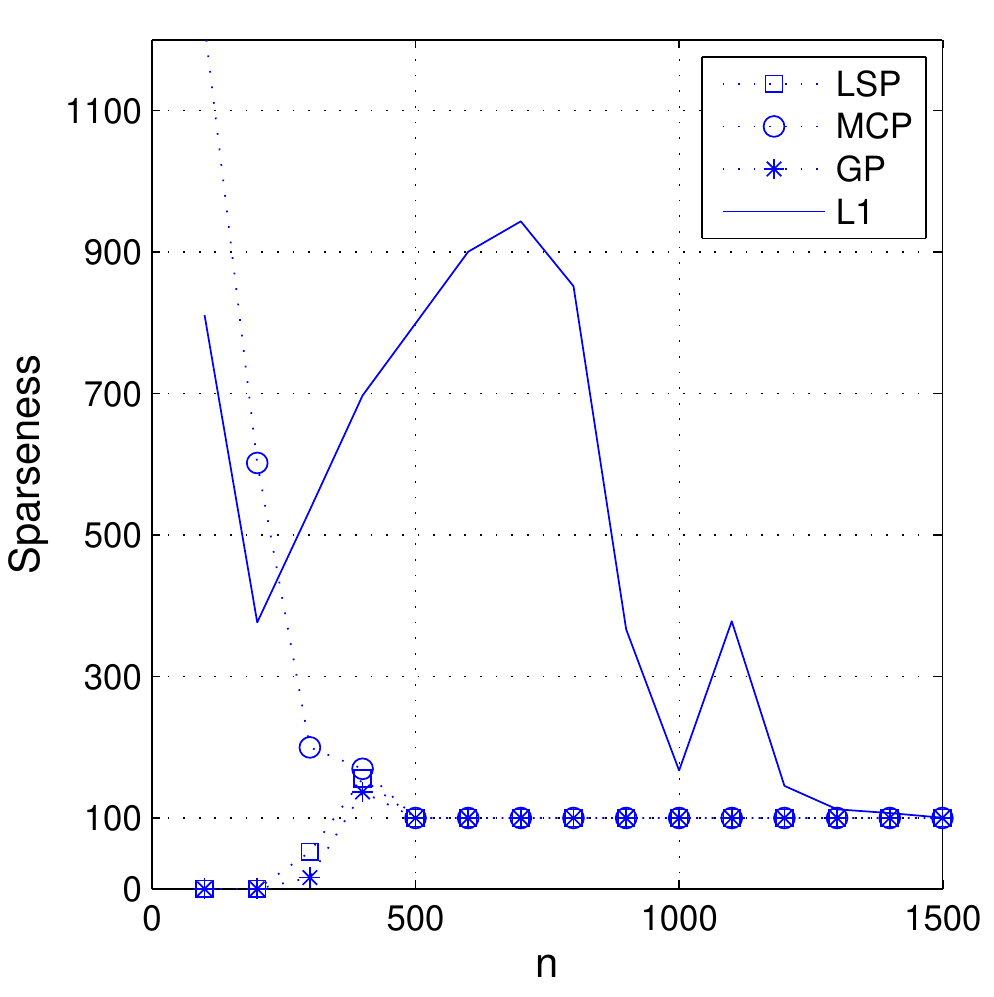}
  \includegraphics[width=0.32\textwidth]{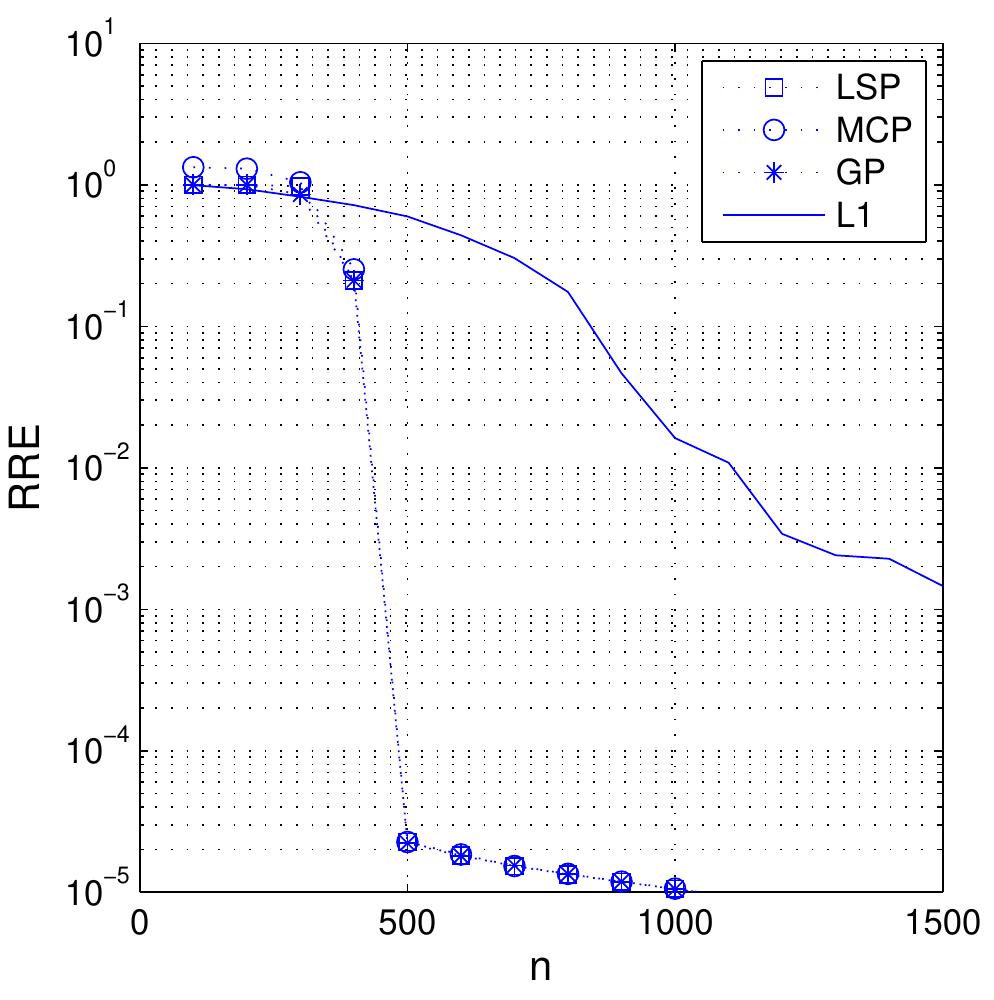}
  \includegraphics[width=0.32\textwidth]{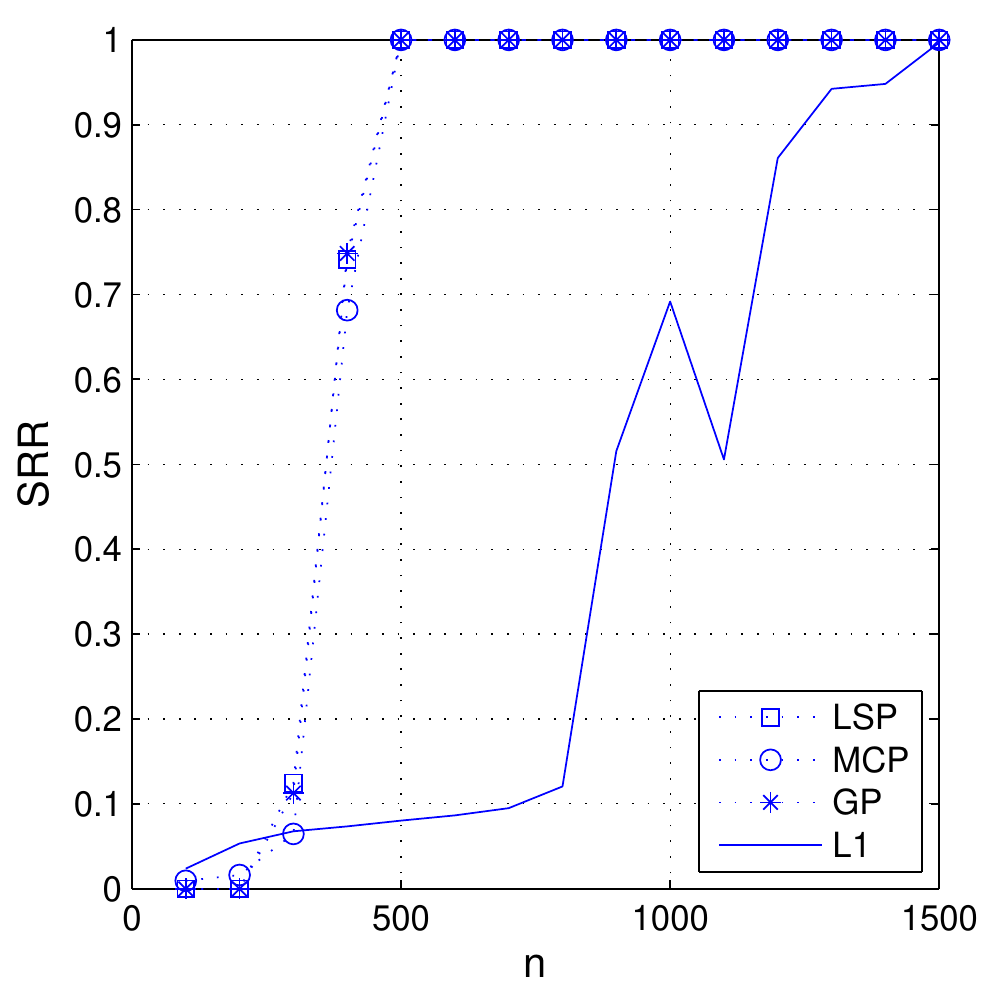}\\
  \caption{The sparseness (left), RRE (middle) and SRR (right) corresponding to the regularizers(LSP, MCP, GP and $\ell_1$-norm). The true parameters, the design matrices and the noises are generated in the same way as Section \ref{sec:10} except that $p=10~000$, $s=100$ and $n$ varies from $s$ to $15s$. The parameter of the regularizers $\gamma$ is set as $10^{-7}$. We use the OMP \cite{tropp2007signal} to generate an initial solution for CD with at most $(n-s)$ non-zero components. The parameters of CD $\psi = 0.1$ and the stopping criterion of CD is the same as Section \ref{sec:10}. Every data point is the average of 100 trials of CD methods. For each regularizer and each $n$, we select $\lambda$ from $10^{-6},~ 10^{-5}, \cdots,~ 10$ such that it gets the smallest average RRE of the 100 trials.}\label{fig:8}
\end{figure}

We fix the dimension of the parameters and the sparseness of the true parameters and we vary the sampling size $n$ to compare the three criterions between convex regularizers ($\ell_1$-norm, implemented by FISTA \cite{beck2009fast}) and non-convex regularizers (LSP, MCP and GP).As Figure \ref{fig:8} shows, non-convex regularizers give much more accurate sparseness estimation, lower RREs and higher SRRs than $\ell_1$-regularization. Among the three non-convex regularizers, the performance of sparse estimation is similar to each other.

\subsection{Single-Pixel Camera}

\begin{figure}
  \centering
  \subfigure[Original]
  {\includegraphics[width=0.2\textwidth]{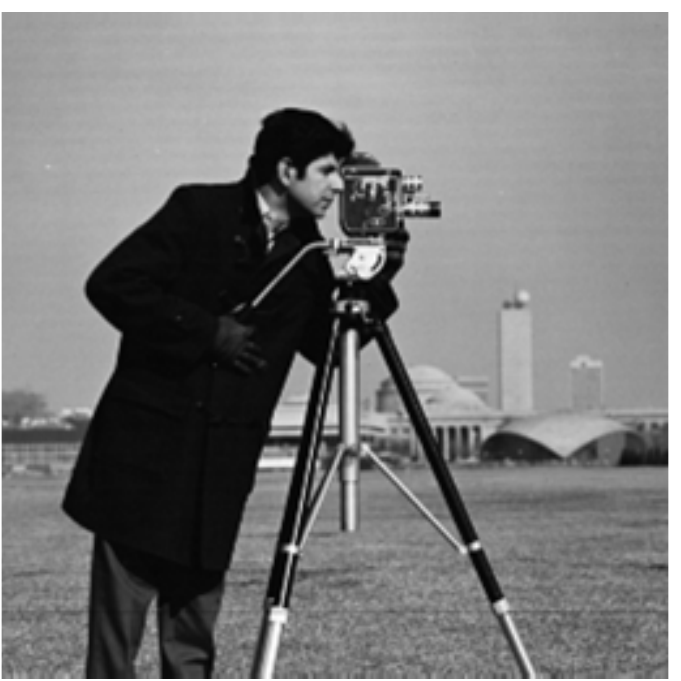}}
  \subfigure[LSP]
  {\includegraphics[width=0.2\textwidth]{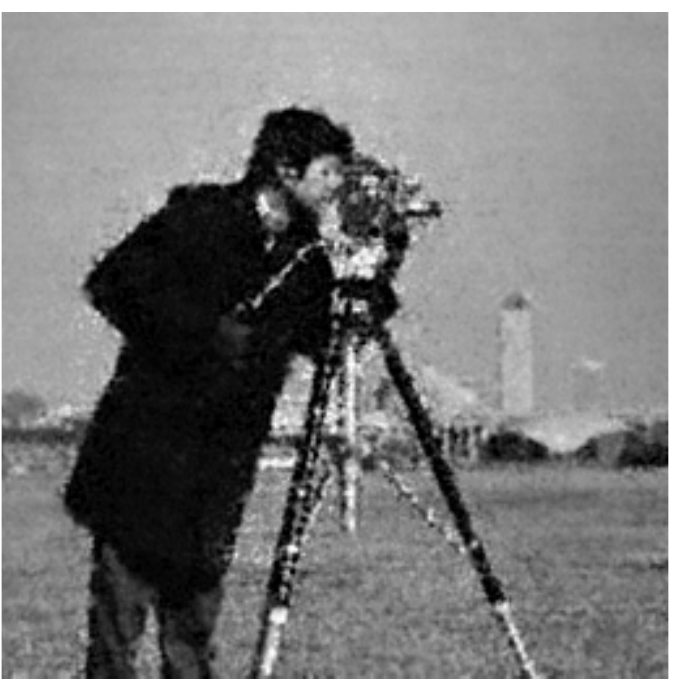}}
  \subfigure[$\ell_1$-norm]
  {\includegraphics[width=0.2\textwidth]{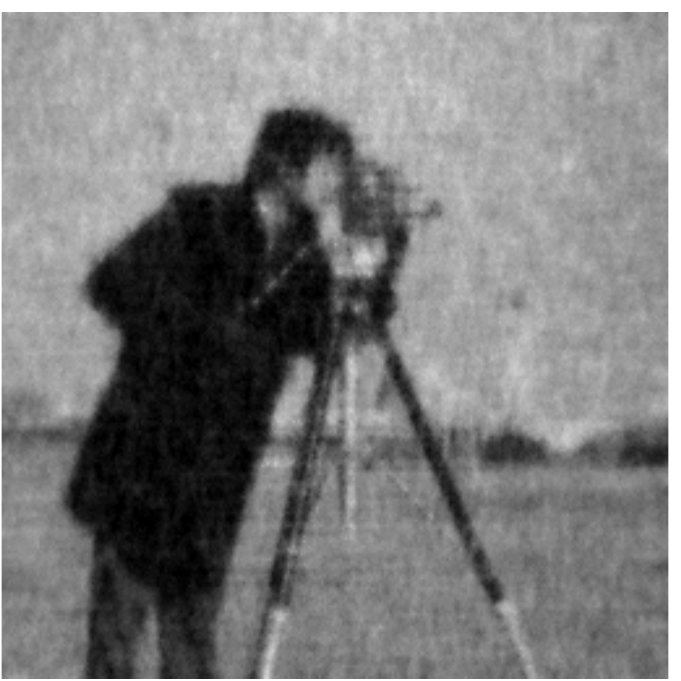}}
  \subfigure[PSNR]
  {\includegraphics[width=0.31\textwidth]{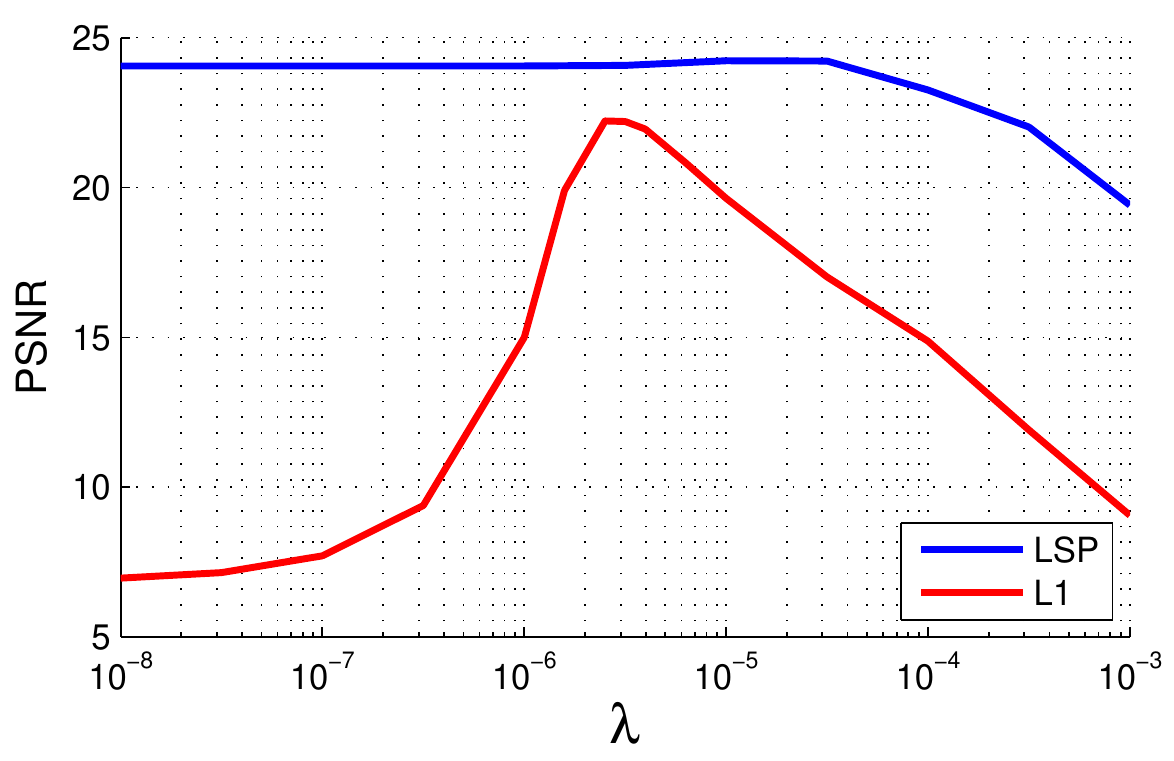}}\\
  \caption{Comparison of image recovery. (a) The original image. (b)(c) The estimated image by LSP and $\ell_1$-norm with highest PSNRs in (d). (d) The PSNRs of LSP and $\ell_1$-norm for different values of $\lambda$. The results of LSP and $\ell_1$-norm are obtained by CD ($\psi=0.001\xi$, $\theta^{(0)} =0$) and FISTA respectively.}\label{fig:2}
\end{figure}

We compare non-convex regularizers and $\ell_1$-norm in the application of single-pixel camera \cite{duarte2008single}. In this application, we need to recover an image from a small fraction of pixels of an image, which is a similar task to image inpainting \cite{mairal2008sparse}. Since most of natural images have sparse Discrete Cosine Transformations~(DCT), we can recover the image by solving the problem $\min_\theta \|y-M \text{vec}(\theta)\|_2^2/(2n) + \mathcal R(\text{vec}(D[\theta]))$, where $y$'s components are the known pixels, $\theta$ is the estimated image, $M$ is a mask matrix indicating the positions of the known pixels, $D[\theta]$ is the 2D-DCT of $\theta$ and $\text{vec}(\theta)$ is the vectorization of $\theta$. Denote $\Theta = D[\theta]$ and we rewrite the problem in the form of Problem (\ref{eqn:1}) $\min_\Theta \|y-M \text{vec}(D^{-1} [\Theta] ) \|_2^2/(2n) + \mathcal R(\text{vec}(\Theta))$, where $D^{-1}[\Theta]$ is the inverse 2D-DCT of $\Theta$.
Figure \ref{fig:2}(a) shows the test image (size 256$\times$256). We randomly choose 25\% pixels of it as $y$. The PSNRs of LSP ($\gamma=10^{-7}$) and $\ell_1$-norm are compared in Figure \ref{fig:2}(d), where LSP has higher PSNRs than $\ell_1$-norm for all $\lambda$s in the figure. The PSNRs of LSP are more robust to $\lambda$ than $\ell_1$-norm. Figure \ref{fig:2}(b) and (c) illustrate the recovered images by LSP and $\ell_1$-norm with the best PSNRs. The image produced by LSP is of better quality than the one created by $\ell_1$-norm.

\section{Conclusion}
This paper establishes a theory for sparse estimation with non-convex regularized regression. The framework of non-convex regularizers in this paper is general and especially suitable for sharp concave regularizers. For proper sharp concave regularizers, both global solutions and AGAS solutions can give good parameter estimation and sparseness estimation. The proposed SE based estimation conditions are weaker than that of $\ell_1$-norm. To obtain AGAS solutions, we give a prediction error based guarantee for AG property and prove that CD methods yield the desired AGAS solutions.

Our theory explains the improvements on sparse estimation from $\ell_1$-regularization to non-convex regularization. Our work can serve as a guideline for the further study on designing regularizers and developing algorithms for non-convex regularization.

\section{Technical Proofs}

We first provide two lemmas. The first is Lemma 1 of \citet{zhang2011general}.

\begin{lemma}\label{lem:9}\hint{lem:9}
Let $\hat\theta$ be a global optima of Problem (\ref{eqn:1}). We have
\hint{eqn:11}
\begin{equation}\label{eqn:11}
\|X^T (X\hat \theta - y)/n\|_\infty \le \lambda^*.
\end{equation}
Under the $\eta$-null consistency condition, we further have
\hint{eqn:19}
\begin{equation}\label{eqn:19}
\|X^T e /n\|_\infty \le \eta \lambda^*.
\end{equation}
\end{lemma}
\begin{lemma}\label{lem:8}
\hint{lem:8}
\begin{enumerate}
  \item $r(u)$ is subadditive, i.e., $r(u_1+u_2) \le r(u_1) + r(u_2), ~ \forall u_1, u_2 \ge 0.$
  \item For any $\forall u > 0$ and any $d \in \partial r(u)$,  $\dot r(0+) \ge \dot r(u-) \ge d \ge \dot r(u+) \ge 0.$
\end{enumerate}
\end{lemma}

\textbf{Proof.} 1. Since $r(u)$ is concave, it follows that $\forall u_1,u_2 \ge 0$, $\frac{ u_1}{ u_1 +  u_2} r( u_1+ u_2) + \frac{ u_2}{ u_1 +  u_2} r(0) \le r( u_1)$ and $ \frac{u_2}{ u_1 +  u_2} r(u_1 + u_2) + \frac{u_1}{u_2 + u_2} r(0) \le r(u_2)$. Summing up the two inequalities gives $r(u_1 + u_2) \le r(u_1) + r(u_2)$.

2. Invoking the subadditivity, we have $[r(u-\Delta u) - r(u)]/\Delta u \le r(\Delta u)/\Delta u$ for $\Delta u>0 $ and $u \ge \Delta u$. Let $\Delta u \to 0$. Then $\dot r(0+) \ge \dot r(u-)$.

The concavity of $r(u)$ yields that $\frac{r(u) - r(u-\Delta u)}{\Delta u} \ge \frac{r(u + \Delta u) - r(u)}{ \Delta u}$ for $\Delta u>0$. From the definition of subgradient of concave function, we have $\Delta u \cdot d \ge r(u+\Delta u) - r(u)$ and $-\Delta u \cdot d \ge r(u - \Delta u) - r(u)$ for any $\Delta u>0$. Hence, $\frac{r(u) - r(u-\Delta u)}{\Delta u} \ge d\ge \frac{r(u + \Delta u) - r(u)}{ \Delta u}$. Let $\Delta u \to 0$ and then the lemma follows. $\blacksquare$

\subsection{Sharp concavity and strong concavity}\label{sec:14}
\hint{sec:14}
Invoking Eqn. (\ref{eqn:57}) with $\alpha>0$, $t_1=0$ and $t_2= t>0$, we have $r((1-\alpha)t) \ge (1-\alpha)r(t) + C\alpha (1-\alpha)t^2/2$, which implies
\[
r(t) \ge t \cdot \frac{r(t) - r(t-\alpha t)}{\alpha t} + C(1-\alpha)t^2/2.
\]
Let $\alpha \to 0$. Sharp concavity follows.

\subsection{The upper bound of $\lambda^*$ for LSP}\label{sec:13}\hint{sec:13}

Define $U>0$ such that $\frac{U^2}{\log(1+U)} = \frac{2}{\xi \gamma^2}$. Let $u=U\lambda \gamma$ and we have $\lambda^* \le \lambda (\frac{\xi \gamma U}{2} + \frac{\log(1+U)}{\gamma U}) = \lambda \sqrt{2 \xi \log(1+U)}$. Note that $U \le \frac{U^2}{\log(1+U)} = \frac{2}{\xi \gamma^2}$. Hence, $\lambda^* \le \lambda \sqrt{2 \xi \log(1+\frac{2}{\xi \gamma^2})}$.
Also, $a_\gamma \le \sqrt{2 \xi \log(1+\frac{2}{\xi \gamma^2})}$.

\subsection{Proof of Theorem \ref{thm:4}}

$\hat\theta$ minimizes $\frac{1}{2n}\|y-X\theta\|_2^2 + \mathcal R(\theta)$, therefore the subgradient at $\hat \theta$ contains zero, i.e., $ |x_i^T(X\hat \theta - y)/n| \le \dot r(|\hat\theta_i|-)$ for any $i\in \mathrm{supp}(\hat\theta)$. Define $\bar \theta = (\hat \theta_1, \cdots,\hat \theta_{i-1}, 0 ,\hat \theta_{i+1} ,\cdots,\hat \theta_n) $. We have $\frac{1}{2n} \|y-X\hat \theta\|_2^2 + \mathcal R(\hat \theta) \le \frac{1}{2n} \|y-X \bar \theta\|_2^2 + \mathcal R(\bar \theta)$, which implies $2n r(|\hat \theta_i|) \le \hat \theta_i^2 \|x_i\|_2^2 + 2 \hat \theta_i x_i^T (y-X\hat \theta) \le \hat \theta_i^2 \|x_i\|_2^2 + 2 |\hat \theta_i| | x_i^T (y-X\hat \theta)| \le n \xi \hat \theta_i^2 + 2n |\hat \theta_i| \dot r(|\hat\theta_i|-)$. If $\hat \theta_i \in (0,u_0)$, this inequality contradicts with $\xi$-sharp concavity condition. $\blacksquare$

\subsection{Proof of Theorem \ref{thm:10}}
We assume that $\theta=0$ is not a minimizer of $\min_{\theta} \frac{1}{2n} \|X \theta - e/\eta\|_2^2 + \mathcal R(\theta)$ while $\hat\theta_\eta\not=0$ is a minimizer. Therefore, $\frac{1}{2n\eta^2} \|e\|_2^2 > \frac{1}{2n}\|X \hat \theta_\eta - e/\eta\|_2^2 + \mathcal R(\hat\theta_\eta)$. Since $r(u)$ is $\xi$-sharp concave over $(0,u_0)$, the non-zero components of $\hat\theta_\eta$ has magnitudes larger than $u_0$. Thus, $\frac{1}{2n}\|X \hat \theta_\eta - e/\eta\|_2^2 + \mathcal R(\hat\theta_\eta) \ge r(u_0) \ge \frac{1}{2n\eta^2 }\|e\|_2^2$. It contradicts with the assumption. $\blacksquare$

\subsection{Proof of Theorem \ref{thm:1}}\label{sec:9}\hint{sec:9}

Let $\Delta=\hat \theta - \theta^* $, $\mathcal S = \mathrm{supp}(\theta^*)$, $s=|\mathcal S|$ and $\mathcal T$ be any index set with $|\mathcal T| \le s$. Let $i_1, i_2, \cdots$ be a sequence of indices such that $i_k \in \bar{\mathcal T}$ for $k \ge 1$ and $|\Delta_{i_1}| \ge |\Delta_{i_2}| \ge |\Delta_{i_3}| \ge \cdots$. Given an integer $t \ge s$, we partition $\bar{\mathcal T}$ as $\bar{\mathcal T} = \cup_{i\ge 1}\mathcal T_i$ such that $\mathcal T_1 = \{i_1, \cdots, i_t\}$, $\mathcal T_2 = \{ i_{t+1}, \cdots, i_{2t}\}$,~$\cdots$. Define $\Sigma = \sum_{i\ge 2} \|\Delta_{\mathcal T_i}\|_2$, $\alpha = (1+\eta) / (1-\eta)$. Before the proof, we introduce the following three lemmas. Lemma \ref{lem:3} is a special case of Lemma \ref{lem:7} with $\mu = 0$.

\begin{lemma}\label{lem:3}\hint{lem:3}
Under $\eta$-null consistency, $\frac{1}{2n} \|X \Delta\|_2^2 + \mathcal R(\Delta_{\bar{\mathcal S}}) \le  \alpha \mathcal R(\Delta_{\mathcal S})$.
\end{lemma}

\begin{lemma}\label{lem:2}\hint{lem:2}
$r(\Sigma/\sqrt{t}) \le \mathcal R(\Delta_{\bar{\mathcal T}}) / t.$
\end{lemma}

\textbf{Proof.}
For any $i\in \mathcal T_k$ and  $j\in \mathcal T_{k-1}$ ($k \ge 2$), we have $|\Delta_i| \le |\Delta_j|$. Thus, $r(|\Delta_i|) \le \mathcal R(\Delta_{\mathcal T_{k-1}})/t$, i.e., $|\Delta_i|^2 \le ( r^{-1}(\mathcal R(\Delta_{\mathcal T_{k-1}})/t))^2$. It follows that $r( \|\Delta_{\mathcal T_k}\|_2 / \sqrt{t} ) \le \mathcal R(\Delta_{\mathcal T_{k-1}})/t$. Thus, $\mathcal R(\Delta_{\bar{\mathcal T}})/ t \ge \sum_{k \ge 2} \mathcal R(\Delta_{\mathcal T_{k-1}})/t \ge \sum_{k \ge 2} r( \|\Delta_{\mathcal T_k}\|_2 / \sqrt{t} ) \ge r(\Sigma/\sqrt{t}). ~\blacksquare $

\begin{lemma}\label{lem:1}
\hint{lem:1}
Under $\eta$-null consistency,
\hint{eqn:3}
\begin{equation}\label{eqn:3}
\max\{ \|\Delta_{\mathcal T}\|_2, \|\Delta_{\mathcal T_1}\|_2 \} \le \frac{1+\sqrt{2}}{2 \kappa_-(2t)} \left[ \frac{\kappa_+(2t) - \kappa_-(2t)}{2} \Sigma + \sqrt{t} (1+\eta)\lambda^* \right].
\end{equation}
\end{lemma}

\textbf{Proof.}
By Lemma \ref{lem:9}, we have $\| X^T X \Delta/n\|_\infty \le  \|X^T(X \hat \theta - y)/n\|_\infty  +  \| X^T e/n\|_\infty \le \lambda^* + \eta \lambda^*$. We modify the Eqn. (12) in \citet{foucart2009sparsest} to the following inequality.
\[
\frac{1}{n} \left< X \Delta,X( \Delta_{\mathcal T} + \Delta_{\mathcal T_1}) \right>
\le ( \|\Delta_{\mathcal T}\|_1 + \|\Delta_{\mathcal T_1}\|_1 ) \|\frac{1}{n} X^T X \Delta\|_\infty
\le \sqrt{t} (1+\eta) \lambda^* (\|\Delta_{\mathcal T}\|_2 + \|\Delta_{\mathcal T_1}\|_2).
\]
Then, following the proof of Theorem 3.1 in \citet{foucart2009sparsest}, Eqn. (\ref{eqn:3}) follows.$\blacksquare$

Next, we turn to the proof Theorem \ref{thm:1}. Let $\kappa_- = \kappa_-(2t)$, $\kappa_+ = \kappa_+(2t)$, $H_r = H_r(\rho_0,\alpha,s,t)$ and $\varrho = (1+\sqrt{2}) (\kappa_+ / \kappa_- -1) /4 $. There are two cases according to the difference of supports of $\hat \theta$ and $\theta^*$.

Case 1: $\text{supp}(\hat \theta) = \text{supp}(\theta^*)$. For this case, we have $\Delta_i=0$ for $i\in \bar{\mathcal S}$ and $\Sigma = 0$, with which and Lemma \ref{lem:1}, we obtain that $\|\Delta\|_2 = \|\Delta_{\mathcal S}\|_2 \le c_1 \lambda^*,$ where $c_1 = (1+\sqrt{2}) (1+\eta) \sqrt{t} / (2\kappa_-) $.

Case 2: $\text{supp}(\hat \theta) \not= \text{supp}(\theta^*)$. Let $\mathcal T$ be the indices of the first $s$ largest components of $\Delta$ in the sense of magnitudes. From the concavity of $r(u)$, $\mathcal R(\Delta_{\mathcal T}) \le s r(\|\Delta_{\mathcal T}\|_1/s) \le s r(\|\Delta_{\mathcal T}\|_2/\sqrt{s})$. By Lemma \ref{lem:1}, we have
\hint{eqn:17}
\begin{equation}\label{eqn:17}
\mathcal R(\Delta_{\mathcal T})
\le s r \left( \frac{\|\Delta_{\mathcal T}\|_2}{\sqrt{s}} \right)
\le s r \left( \frac{1+\sqrt{2}}{2\sqrt{s} \kappa_- } \left( \frac{\kappa_+ - \kappa_-}{2} \Sigma + \sqrt{t}(1+\eta) \lambda^*\right) \right).
\end{equation}
Combining with Lemma \ref{lem:3} and \ref{lem:2}, it follows that
\hint{eqn:35}
\begin{equation}\label{eqn:35}
r^{-1}\left( \frac{ \mathcal R(\Delta_{\mathcal T})}{s}\right)
- \varrho \sqrt{\frac{t}{s}}
r^{-1}\left( \frac{ \alpha \mathcal R(\Delta_{\mathcal T})}{t} \right)
\le
\frac{(1+\sqrt{2}) (1+\eta)}{2 \kappa_-} \sqrt{\frac{t}{s}} \lambda^*.
\end{equation}
By the definition of $\rho_0$ in Eqn. (\ref{eqn:60}) and $\text{supp}(\hat \theta) \not= \text{supp}(\theta^*)$, there exists $j$ satisfying $|\Delta_j| \ge \rho_0$, which implies $\mathcal R(\Delta_{\mathcal T}) \ge r(\rho_0)$. Since $\frac{r^{-1}(u/s)}{r^{-1}(\alpha u/t)}$ is a non-decreasing function of $u$, we have that
\[
\frac{r^{-1}(\mathcal R(\Delta_{\mathcal T})/s)}{r^{-1}( \alpha \mathcal R( \Delta_{\mathcal T})/t)} \ge \frac{r^{-1}(r(\rho_0)/s)}{r^{-1}(\alpha r(\rho_0)/t)} = \sqrt{\frac{t}{s}} H_r(\rho_0,\alpha,s,t).
\]
for $\rho_0>0$. If $\rho_0=0$, the left hand of the above inequality still holds since $H_r(0,\alpha,s,t) = \lim_{\rho\to 0+} H_r(\rho,\alpha,s,t)$. Under the condition $H_r - \varrho > 0$, we have
\hint{eqn:40}
\begin{equation}\label{eqn:40}
r^{-1}\left( \alpha \mathcal R(\Delta_{\mathcal S}) / t \right) \le r^{-1}\left( \alpha \mathcal R(\Delta_{\mathcal T}) / t \right) \le C_2 (1+\eta) \lambda^*,
\end{equation}
where
\hint{eqn:13}
\begin{equation}\label{eqn:13}
C_2 = \frac{1+\sqrt{2}}{2 (H_r - \varrho)\kappa_- } .
\end{equation}
Hence, we have $\Sigma \le \sqrt{t} C_2 (1+\eta) \lambda^*$ by Lemma \ref{lem:3} and Lemma \ref{lem:2}. Invoking Lemma \ref{lem:1} and $\|\Delta\|_2
\le \|\Delta_{\mathcal T}\|_2 + \|\Delta_{\mathcal T_1}\|_2 + \Sigma$, the conclusion follows with some algebra. $\blacksquare$

\subsection{Proof of Theorem \ref{thm:8}}
The proof is similar to Theorem 2 in \citet{zhang2011general} except that we bound $\mathcal R(\Delta_{\mathcal S})$ and $\|X \Delta\|_2^2/(2n)$ as follows. By Eqn. (\ref{eqn:40}), we have $\mathcal R(\Delta_{\mathcal S}) \le \frac{t}{\alpha} r(C_2 (1+\eta)\lambda^* )$ and $\frac{1}{2n} \|X \Delta\|_2^2 \le \alpha \mathcal R(\Delta_{\mathcal S}) \le t r(C_2 (1+\eta) \lambda^*)$.

\subsection{The method to obtain Eqn. (\ref{eqn:26}) and (\ref{eqn:16})}\label{sec:8}
\hint{sec:8}

Suppose $r(u) = C u^q~(0<q\le 1)$  for $u\ge \lambda \gamma (1-\phi)$. The continuity and the concavity of $r(u)$ require that $C (\lambda \gamma(1-\phi))^q = 0.5\lambda^2 \gamma(1-\phi^2)$ and $Cq (\lambda \gamma (1-\phi))^{q-1} \le \lambda \phi$. Thus, it is feasible that $q=2\phi/(1+\phi)$ and $C = 0.5 \lambda^2 \gamma (1-\phi^2)/ (\lambda \gamma (1-\phi))^q$. Eqn. (\ref{eqn:26}) follows. For this setting for $C$ and $q$, $r(u)$ is $\xi$-sharp concave over $(0,\rho_0)$ with $\rho_0 = \lambda \gamma (1-\phi) (\frac{\phi}{\xi\gamma (1+\phi)})^{(1+\phi)/2}$. We observe that $r(\rho_0)/s \ge \lambda^2 \gamma (1-\phi^2)/2 = \alpha r(\rho_0)/t$ holds under the condition that $\frac{\alpha}{t} (\frac{\phi}{\gamma \xi (1+\phi)})^\phi =1$, i.e., $\gamma \xi= \frac{\phi}{1+\phi} (\alpha/t)^{1/\phi}$. Thus, $r^{-1} (\alpha r(\rho_0) / t ) = \lambda \gamma (1-\phi)$ and $r^{-1} ( r(\rho_0)/s) = \lambda \gamma (1-\phi) (t/(\alpha s))^{1/q}$ with $q=2\phi/(1+\phi)$. Then, Eqn. (\ref{eqn:16}) follows.

\subsection{Proof of Theorem \ref{thm:9}}

Let $\Delta = \hat \theta - \theta^*$. By Lemma \ref{lem:3} in Section \ref{sec:9}, we have $\text{RE}^{\mathcal R}(\alpha, \mathcal S) \|\Delta\|_2^2 \le \|X \Delta\|_2^2/n$ and $\text{RIF}^{\mathcal R}_\tau (\alpha, \mathcal S) \|\Delta\|_\tau \le s^{1/\tau} \|X^T X \Delta\|_\infty/n$.Invoking null consistency, we have $ e^T X\Delta/n \le \eta\|X \Delta\|_2^2/(2n) + \eta \mathcal R(\Delta)$. Then,
\[
\begin{array}{ll}
0 & \ge \mathcal L(\theta^* + \Delta) - \mathcal L(\theta^*) + \mathcal R(\theta^* + \Delta) - \mathcal R(\theta^*) \\
& \ge \|X \Delta\|_2^2/(2n) - e^T X \Delta/n + \mathcal R(\Delta_{\bar{\mathcal S}}) - \mathcal R(\Delta_{\mathcal S}) \\
& \ge (1-\eta) \|X \Delta\|_2^2/(2n) - (1+\eta) \mathcal R(\Delta)  \\
& \ge (1-\eta) \|\Delta\|_2^2 \text{RE}^{\mathcal R}(\alpha, \mathcal S)/2 - (1+\eta) \sqrt{s} \dot r(0+) \|\Delta\|_2.
\end{array}
\]

Hence, we obtain $\|\Delta\|_2 \le \frac{ 2\alpha\sqrt{s}}{ \text{RE}^{\mathcal R} (\alpha, \mathcal S) } \dot r(0+)$. By Lemma \ref{lem:9}, $\|X^T X \Delta/n\|_\infty \le \|X^T (X \hat \theta -y)/n\|_\infty + \|X^T e/n\|_\infty \le (1+\eta)\lambda^*$. By the definition of RIF, we have $\|\Delta\|_\tau \le \frac{(1+\eta)\lambda^* s^{1/\tau} }{\text{RIF}^{\mathcal R}_\tau (\alpha, \mathcal S)}$. $\blacksquare$

\subsection{Proof of Theorem \ref{thm:3}}

The proof needs the following two lemmas, which are extensions of Lemma \ref{lem:3} and Lemma \ref{lem:1}  The notations are the same as Section \ref{sec:9} except that $\Delta = \tilde\theta - \theta^*$.

\begin{lemma}\label{lem:7}\hint{lem:7}
Suppose $\tilde \theta$ is a $(\theta^*, \mu)$-approximate global solution and the regularized regression satisfies the $\eta$-null consistency condition. Then, $\|X \Delta \|_2^2/(2n) + \mathcal R(\Delta_{\bar{\mathcal S}}) \le  \alpha \mathcal R(\Delta_{\mathcal S}) + \mu/(1-\eta)$.
\end{lemma}

\textbf{Proof.}
Invoking $\eta$-null consistency condition, we have $e^T X\Delta/n \le \eta \|X \Delta\|_2^2/(2n) + \eta \mathcal R(\Delta)$. Since $\tilde\theta = \theta^*+ \Delta $ is a $(\theta^*, \mu)$-approximate global solution, we have
\[
\begin{array}{ll}
\mu & \ge \mathcal L(\theta^*+\Delta) - \mathcal L(\theta^*) + \mathcal R(\theta^*+\Delta) - \mathcal R(\theta^*)\\
& \ge \|X\Delta\|_2^2/(2n) - e^T X\Delta/n + \mathcal R(\Delta_{\bar{\mathcal S}}) - \mathcal R(\Delta_{\mathcal S}) \\
& \ge (1-\eta) \|X \Delta\|_2^2/(2n) -\eta \mathcal R(\Delta) + \mathcal R(\Delta_{\bar{\mathcal S}}) - \mathcal R(\Delta_{\mathcal S})
\end{array}
\]
Hence, the conclusion follows. $\blacksquare$

\begin{lemma}\label{lem:6}\hint{lem:6}
Under $\eta$-null consistency,
\hint{eqn:7}
\begin{equation}\label{eqn:7}
\max\{ \|\Delta_{\mathcal T}\|_2, \|\Delta_{\mathcal T_1}\|_2 \} \le \frac{1+\sqrt{2}}{2 \kappa_-(2t)} \left[ \frac{\kappa_+(2t) - \kappa_-(2t)}{2} \Sigma + \sqrt{t} \tilde\epsilon \right].
\end{equation}
\end{lemma}

\textbf{Proof.}
Since $\tilde \theta$ is a $\nu$-AS solution, we have $\| X^T(X \tilde \theta - y)/n \|_\infty \le \dot r(0+) +  \nu$. From the triangle inequality and Eqn. (\ref{eqn:19}), we have $\|X^T X \Delta/n\|_\infty \le  \|X^T(X \tilde \theta - y)/n\|_\infty  + \|X^T e/n\|_\infty \le \dot r(0+) +  \eta \lambda^* + \nu = \tilde\epsilon$. Eqn. (\ref{eqn:7}) follows with the same analysis as the proof of Lemma \ref{lem:1}. $\blacksquare$

Next, we turn to the proof of Theorem \ref{thm:3}. The proof is similar to that of Theorem \ref{thm:1}. Here we only provide some important steps. Let $\kappa_- = \kappa_-(2t)$, $\kappa_+ = \kappa_+(2t)$, $G_r = G_r(\tilde\rho_0,\alpha,s,t)$ and $\varrho = (1+\sqrt{2}) (\kappa_+ / \kappa_- -1) /4 $.
%There are two cases according to the difference of supports of $\hat \theta$ and $\theta^*$.

Case 1: $\text{supp}(\tilde \theta) = \text{supp}(\theta^*)$. Similar to Case 1 of Theorem \ref{thm:1}, we have $\|\Delta\|_2 = \|\Delta_{\mathcal S}\|_2 \le c_3 \tilde\epsilon$ where $c_3 = (1+\sqrt{2}) \sqrt{t} / (2\kappa_-)$.

Case 2: $\text{supp}(\tilde \theta) \not= \text{supp} (\theta^*)$.
%For this case, There exists $j$ satisfying $|\Delta_j| \ge \tilde\rho_0$. Let $\mathcal T$ be the indices of the first $s$ largest components of $\Delta$ in the sense of magnitudes. Obviously, $\mathcal R(\Delta_{\mathcal T}) \ge r(\tilde\rho_0)$.
Similar to Eqn. (\ref{eqn:35}), we have
%\hint{eqn:36}
\begin{equation}\label{eqn:36}
r^{-1} \left( \frac{\mathcal R(\Delta_{\mathcal T})}{s} \right) -
\varrho \sqrt{\frac{t}{s}} r^{-1} \left( \frac{\alpha \mathcal R(\Delta_{\mathcal T})}{t} + \frac{\mu}{(1-\eta)t} \right)
\le
\frac{1+\sqrt{2}}{2\kappa_-} \sqrt{\frac{t}{s}} \tilde\epsilon
\end{equation}
Since $r(u)$ is non-decreasing and concave, $r^{-1}(u)$ is convex. Therefore,
%\hint{eqn:37}
\begin{equation}\label{eqn:37}
r^{-1} \left( \frac{\alpha \mathcal R(\Delta_{\mathcal T})}{t} + \frac{\mu}{(1-\eta)t} \right)
\le
\frac{t-1}{t} r^{-1} \left( \frac{\alpha \mathcal R(\Delta_{\mathcal T})}{t-1} \right)
+
\frac{1}{t}  r^{-1} \left( \frac{\mu}{1-\eta}\right).
\end{equation}
We observe that
%\hint{eqn:38}
\begin{equation}\label{eqn:38}
\frac{r^{-1}(\mathcal R(\Delta_{\mathcal T})/s)}{r^{-1}( \alpha \mathcal R(\Delta_{\mathcal T})/(t-1))}
\ge G_r \frac{t-1}{\sqrt{st}}
\end{equation}
Combining Eqn. (\ref{eqn:36})-(\ref{eqn:38}), we know that under the condition of Eqn. (\ref{eqn:6}),
\hint{eqn:50}
\begin{equation}\label{eqn:50}
r^{-1} ( \alpha \mathcal R(\Delta_{\mathcal S})/(t-1)) \le c_4 \tilde\epsilon + c_5 r^{-1}( \mu/(1-\eta))/(t-1),
\end{equation}
where
\hint{eqn:43}
\begin{equation}\label{eqn:43}
c_4 =\frac{t}{t-1} \frac{1+\sqrt{2}}{2 (G_r - \varrho) \kappa_- }
\end{equation}
and
\hint{eqn:44}
\begin{equation}\label{eqn:44}
c_5 = \varrho/(G_r-\varrho).
\end{equation}
Hence, we have $\Sigma \le \sqrt{t}c_4 \tilde\epsilon + \frac{c_5+1}{\sqrt{t}} r^{-1} \left( \frac{\mu}{1-\eta}\right)$. With this and Lemma \ref{lem:6}, it follows that $\|\Delta\|_2 \le C_4 \tilde\epsilon + C_5 r^{-1} \left( \frac{\mu}{1-\eta}\right)$,
where
\hint{eqn:41}
\begin{equation}\label{eqn:41}
C_4 = \frac{\sqrt{t}(1+\sqrt{2})}{\kappa_-} \frac{G_r + \varrho/(t-1) + 0.5t/(t-1)}{ G_r - \varrho} \ge c_3,
\end{equation}
\hint{eqn:42}
\begin{equation}\label{eqn:42}
C_5 =  \frac{ (2\varrho + 1) G_r}{\sqrt{t} ( G_r - \varrho) }. ~\blacksquare
\end{equation}

\subsection{Proof of Theorem \ref{thm:11}}

Let $\Delta^0= \theta^0 - \theta^*$. We have $\|X \Delta^0\|_2 \le \|X\Delta^0 - e\|_2 + \epsilon \le \mu_0 \sqrt{2n} +\epsilon$. So,
\[
\begin{array}{ll}
\mu & = \mathcal L(\theta^0) -\mathcal L(\theta^*) + \mathcal R(\theta^*+ \Delta^0) -\mathcal R(\theta^*) \\
& \le \mathcal L(\theta^0) + \mathcal R(\Delta^0) \\
& \le \mu_0^2 + (s+s_0) r(\|\Delta^0\|_2/\sqrt{s+s_0}) \\
& \le \mu_0^2 + (s+s_0) r(\|X \Delta^0\|_2/\sqrt{n \kappa_-(s+s_0) (s+s_0}) \\
& \le \mu_0^2 + (s+s_0) r((\epsilon/\sqrt{n}+ \sqrt{2} \mu_0)/\sqrt{(s+s_0) \kappa_-(s+s_0)}) \nonumber. ~\blacksquare
\end{array}
\]

\subsection{Proof of Theorem \ref{thm:6}}
For any $i=1, \cdots, p-1$, let $z_{k,i} = (\theta_1^{(k)}, \cdots, \theta_i^{(k)}, \theta_{i+1}^{(k-1)}, \cdots, \theta_p^{(k-1)})^T$ and $z_{k,0} = \theta^{(k-1)}$, $z_{k,p} = \theta^{(k)}$. By the definition of $\theta_i^{(k)}$ in Eqn (\ref{eqn:22}), we have
\hint{eqn:53}
\begin{equation}\label{eqn:53}
\mathcal F( z_{k, i} ) \le \mathcal F( z_{k, i} ) + \psi ( \theta^{(k)}_{i} -  \theta^{(k-1)}_{i})^2/2 \le \mathcal F( z_{k, i-1} ).
\end{equation}
Thus, $\mathcal F( \theta^{(k)}) = \mathcal F(z_{k,p}) \le \mathcal F( z_{k, i} ) \le \mathcal F( z_{k, i} )+ \psi ( \theta^{(k)}_i -  \theta^{(k-1)}_i)^2/2  \le \mathcal F( z_{k, 0}) = \mathcal F( \theta^{(k-1)})$. Note that $\mathcal F(\theta^{(k)}) \ge 0$ for any $k$. Thus, $\{\mathcal F(\theta^{(k)})\}$, as well as $\{\mathcal F( z_{k, i} )\}$ and $\{\mathcal F( z_{k, i} )+ \psi ( \theta^{(k)}_i -  \theta^{(k-1)}_i)^2/2\} $ are non-increasing sequences and converge to the same non-negative value.

Summing up the right inequality of Eqn. (\ref{eqn:53}) from $i=1$ to $p$, we have $\|\theta^{(k)} - \theta^{(k-1)}\|_2^2 \le 2(\mathcal F(\theta^{(k-1)}) - \mathcal F(\theta^{(k)}) )/\psi$. Summing up from $k=1$ to $K$, we have
\hint{eqn:59}
\begin{equation}\label{eqn:59}
\min_{1\le k \le K} \|\theta^{(k)} - \theta^{(k-1)}\|_2^2 \le \frac{\sum_{k=1}^K \|\theta^{(k)} - \theta^{(k-1)}\|_2^2}{K} \le \frac{2\mathcal F(\theta^{(0)})}{\psi K}
\end{equation}

The directional derivative of Eqn. (\ref{eqn:22}) at $\theta_i^{(k)}$ is non-negative, i.e.,
\hint{eqn:54}
\begin{equation}\label{eqn:54}
d_i x_i^T (Xz_{k,i} - y)/n + \mathcal R'(\theta_i^{(k)};d_i) + \psi(\theta^{(k)}_i - \theta^{(k-1)}_i)d_i \ge 0
\end{equation}
for any $d_i \in \mathbb R$. Summing up Eqn. (\ref{eqn:54}) from $i=1$ to $p$, we have for any $d\in \mathbb R^p$
\begin{equation}
\begin{array}{ll}
0
& \le \sum_{i=1}^p \psi(\theta^{(k)}_i - \theta^{(k-1)}_i)d_i  + \mathcal R'(\theta^{(k)}; d) + \sum_{i=1}^p d_i x_i^T (Xz_{k,i} - y)/n \\
& \le \psi \|d\|_\infty \|\theta^{(k)} - \theta^{(k-1)}\|_1 + \mathcal R'(\theta^{(k)}; d)  \\
& ~~~~~~ + d^T \nabla \mathcal L(\theta^{(k)}) +  \sum_{i=1}^p \sum_{j=i+1}^p d_i (\theta_j^{(k-1)} - \theta_j^{(k)}) x_i^T x_j/n \\
& \le \mathcal F'(\theta^{(k)}; d) + \psi \|d\|_\infty \|\theta^{(k)} - \theta^{(k-1)}\|_1 + \xi \|d\|_\infty \sum_{i=1}^p \sum_{j=i+1}^p |\theta_j^{(k-1)} - \theta_j^{(k)}| \\
& \le \mathcal F'(\theta^{(k)};d) + (\psi + p \xi)\|d\|_\infty \|\theta^{(k)} - \theta^{(k-1)}\|_1 \\
& \le \mathcal F'(\theta^{(k)};d) + (\psi + p \xi)\sqrt{p} \|d\|_\infty \|\theta^{(k)} - \theta^{(k-1)}\|_2
\end{array}
\end{equation}
Hence, $\mathcal F'(\theta^{(k)};d) \ge - (\psi + p \xi)\sqrt{p} \|d\|_\infty \|\theta^{(k)} - \theta^{(k-1)}\|_2$. When CD stops iteration, $\|\theta^{(k)} - \theta^{(k-1)}\|_2 \le \tau = \nu/((\psi + p \xi)\sqrt{p})$ and $\|\theta^{(j)} - \theta^{(j-1)}\|_2 \ge \tau$ for $j\le k-1$, which implies $\mathcal F'(\theta^{(k)};d) \ge -\nu$ for any $\|d\|_2=1$. Invoking Eqn. (\ref{eqn:59}), we have $\tau^2 \le 2\mathcal F(\theta^{(0)})/(\psi(k-1))$. Thus, $k\le 2p (\psi + p\xi)^2 \mathcal F(\theta^{(0)})/(\psi \nu^2) +1 $. $\blacksquare$

\bibliographystyle{plainnat}
\bibliography{papers}
\end{document}